\documentclass[lettersize,journal]{IEEEtran}
\usepackage{amsmath,amsfonts}
\usepackage{algorithmic}
\usepackage{algorithm}
\usepackage{array}
\usepackage[caption=false,font=normalsize,labelfont=sf,textfont=sf]{subfig}
\usepackage{textcomp}
\usepackage{stfloats}
\usepackage{url}
\usepackage{indentfirst}
\usepackage{xcolor}
\usepackage{bm}
\usepackage{verbatim}
\usepackage{graphicx}
\usepackage{cite}

\usepackage{color}
\usepackage{multirow}
\usepackage{multicol}
\usepackage{array}
\usepackage{longtable}
\usepackage{bbding} 

\usepackage{booktabs}
\usepackage{colortbl}  
\usepackage{xcolor}
\usepackage{lipsum}

\usepackage[colorlinks,
            linkcolor=red,
            anchorcolor=blue,
            citecolor=green]{hyperref}
\usepackage{ulem}
\begin{document}
\title{SMFormer: Empowering Self-supervised Stereo Matching via Foundation Models and Data Augmentation}

\author{Yun Wang, Zhengjie Yang*, Jiahao Zheng,  Zhanjie Zhang,  Dapeng Oliver Wu*, Yulan Guo
\thanks{
This work was supported by the InnoHK Initiative, The Government of Hong Kong, SAR (HKSAR), and Laboratory for Artificial Intelligence (AI)-Powered Financial Technologies, and was partially supported by Hong Kong Research Grants Council grant C1042-23GF and Hong Kong Innovation and Technology Commission grant MHP/061/23.

Yun Wang, Jiahao Zheng, and Dapeng Oliver Wu are with the Department of Computer Science, City University of Hong Kong, Kowloon, Hong Kong, China (Email: ywang3875-c@my.cityu.edu.hk, jhzheng4-c@my.cityu.edu.hk, dpwu@ieee.org). 
(Corresponding authors: \textit{Zhengjie Yang} and \textit{Dapeng Oliver Wu})

Zhengjie Yang is with the Hong Kong Generative AI Research and Development Center, The Hong Kong University of Science and Technology, Hong Kong, China (Email: zhengjieyang@ust.hk).

Zhanjie Zhang is with the College of Computer Science and Technology, Zhejiang University, Hangzhou 310000, China (Email: cszzj@zju.edu.cn).

Yulan Guo is with the School of Electronics and Communication Engineering, the Shenzhen Campus of Sun Yat-sen University, Sun Yat-sen University, Shenzhen, China. (Email: guoyulan@sysu.edu.cn). 

}}


\maketitle

\begin{abstract}
Recent self-supervised stereo matching methods have made significant progress. They typically rely on the photometric consistency assumption, which presumes corresponding points across views share the same appearance. However, this assumption could be compromised by real-world disturbances, resulting in invalid supervisory signals and a significant accuracy gap compared to supervised methods.
To address this issue, we propose SMFormer, a framework integrating more reliable self-supervision guided by the Vision Foundation Model (VFM) and data augmentation. We first incorporate the VFM with the Feature Pyramid Network (FPN), providing a discriminative and robust feature representation against disturbance in various scenarios.
We then devise an effective data augmentation mechanism that ensures robustness to various transformations. The data augmentation mechanism explicitly enforces consistency between learned features and those influenced by illumination variations. Additionally, it regularizes the output consistency between disparity predictions of strong augmented samples and those generated from standard samples.
Experiments on multiple mainstream benchmarks demonstrate that our SMFormer achieves state-of-the-art (SOTA) performance among self-supervised methods and even competes on par with supervised ones. 
Remarkably, in the challenging Booster benchmark, SMFormer even outperforms some SOTA supervised methods, such as CFNet.
\end{abstract}


\section{Introduction}

\begin{figure}
    \centering
    \includegraphics[width=1\linewidth]{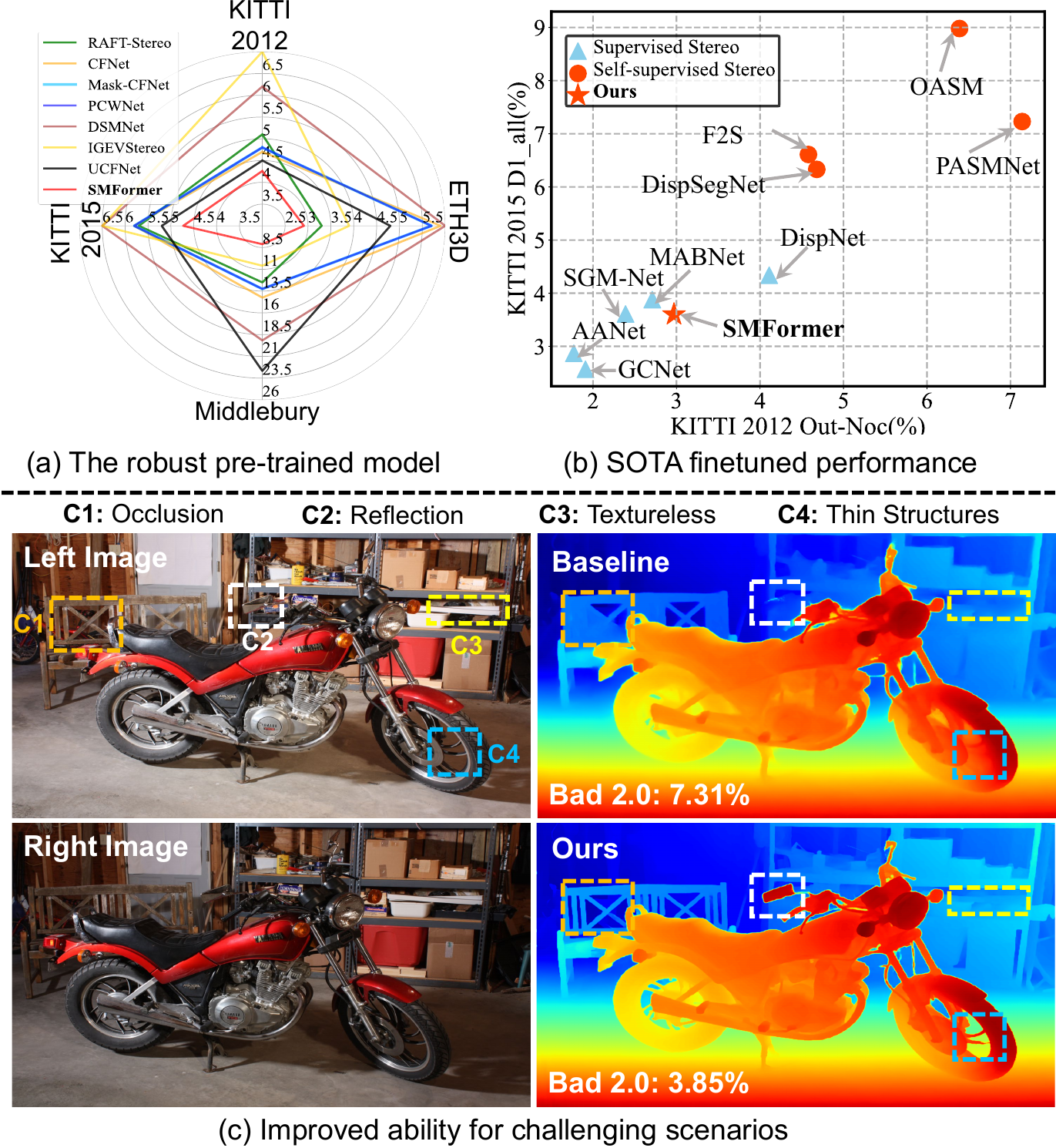}
    \vspace{-0.7cm}
    \caption{(a), our method, equipped with the Vision Foundation Model (SAM~\cite{kirillov2023segment}), which is a robust pre-trained model,
    surpassing previous state-of-the-art domain generalized methods.
    (b), the fine-tuned performance significantly surpasses previous self-supervised methods, even surpassing some supervised ones.
    (c) Our SMFormer, equipped with the proposed feature extractors and self-supervised losses, demonstrates robust performance in challenging regions such as occluded areas (Class 1), reflective areas (Class 2), textureless areas (Class 3), thin structure areas (Class 4), and areas with illumination changes (low lighting in the right image) compared to solely relying on the photometric consistency loss (Baseline). Best zoomed in.} 
    \label{sec1:teaser}
    \vspace{-0.4cm}
\end{figure}

Stereo matching is critical for practical applications, including AR\&VR, robotics, and autonomous driving. 
It aims to find accurate pixel-wise stereo correspondence between rectified stereo pairs.
Existing learning-based   methods~\cite{shen2021cfnet,xu2023iterative,li2022practical,Lipson2021RAFTStereoMR,wang2022spnet,wang2022exploring} typically being cast as an end-to-end disparity/depth regression task, outperform the traditional ones~\cite{hirschmuller2007stereo}.
However, these methods are limited by their dependence on costly ground-truth disparity labels.
Self-supervised stereo matching methods lift such limitations by leveraging the photometric consistency assumption\footnote{The photometric consistency assumes that the appearance of a point in 3D space remains visible and color-invariant across different views.} in place of the costly supervisory signals. 
Despite recent advancements, the assumption of photometric consistency can be disrupted by common real-world disturbances like \textit{reflections, texture-less regions, occlusions}, and \textit{illumination variations}, leading to performance degradation,  as illustrated in Fig.~\ref{sec1:teaser} (c).

In this work, we aim to address the limitations caused by the violation of the photometric consistency assumption, 
incorporating the following additional priors:
(1) The priors of Visual Foundation Models (VFMs) can provide sufficient discriminative features in indistinguishable areas, such as \textit{reflective} and \textit{texture-less} regions.
(2) The priors of data augmentation consistency can enhance the model robustness towards ill-posed regions such as \textit{illumination changes} and \textit{occlusions}.
Hence, we propose a novel self-supervised stereo matching framework, namely SMFormer.

\begin{figure*}
    \includegraphics[width=1.0\linewidth]{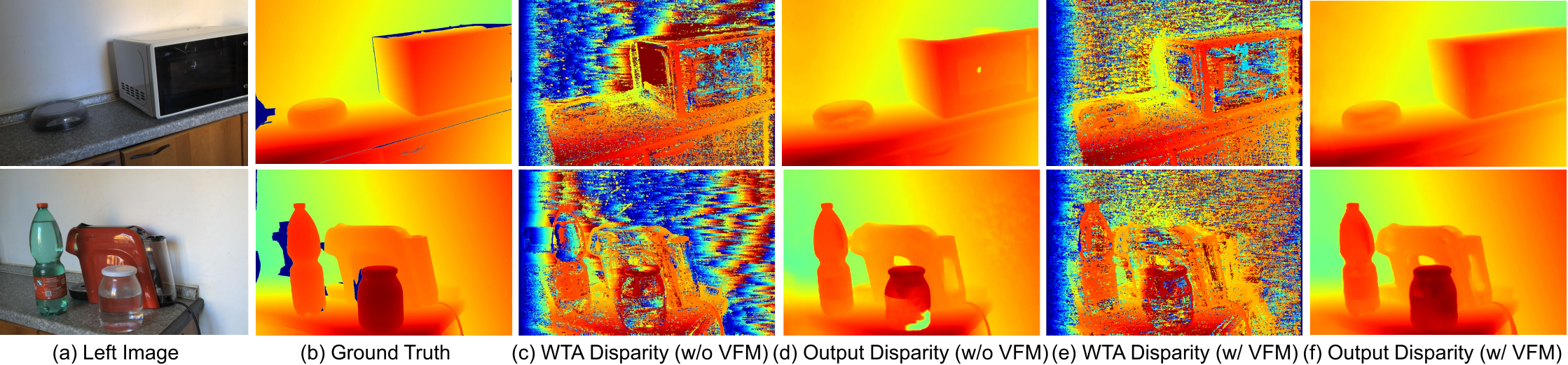}
    \vspace{-0.5cm}
    \caption{Hard cases in Booster~\cite{ramirez2022open} with reflective and texture-less regions. (c), (e) indicates Winner-Take-All (WTA) disparity from the feature correlation (1/4 scale) achieved by the dot products among left and right features before the 3D CNN cost aggregation. WTA disparity enhanced with pre-trained VFM (SAM~\cite{kirillov2023segment}) contains less noise than the original ones without VFM priors.
    Compared with (d), (f) achieves better disparity maps. The models are trained in a self-supervised manner.}
    \vspace{-0.2cm}
    \label{sec1:hard_case}
\end{figure*}

For the prior of VFMs, existing self-supervised stereo matching methods usually leverage CNN-based Feature Pyramid Networks (FPNs)~\cite{wang2020parallax,liu2020flow2stereo,zhang2019dispsegnet} for feature extraction on the target domain, but they struggle with indistinguishable regions like \textit{reflective} and \textit{texture-less} areas due to limited supervisory signals and receptive fields.   
ViT-based VFMs~\cite{oquab2024dinov2,kirillov2023segment,yang2024depth_v2,fang2023eva,dust3r_cvpr24} have demonstrated robustness and generality in learning deep features across various computer vision tasks. 
Several supervised methods~\cite{liu2024playing,zhanglearning2024} develop adapters to retrieve target domain fine-grained feature representations from all-purpose coarse-grained features of the pre-trained VFM. Nevertheless, these methods still struggle to capture rich target domain information, potentially leading to sub-optimal performance~\cite{goyal2023finetune,dutt2023parameter}.
To address this, we propose incorporating robust priors from the pre-trained VFM into FPN features on the target domain, enhancing feature distinctiveness and robustness.
Benefiting from the long-range attention mechanism and multi-scale pyramid architecture, priors from VFM are complementary to those from FPN, enhancing the modeling of challenging areas in self-supervised stereo matching, as shown in Fig.~\ref{sec1:hard_case}.
Meanwhile, since existing VFMs trained for single-image tasks lack sufficient cross-view geometry interaction~\cite{weinzaepfel2023croco}, we design a Multi-layer Attention (MLA) module to introduce cross-view attention across different layers and views, further enhancing the cross-view learning capabilities of VFMs.
For the prior of data augmentation consistency, although data augmentation is critical for supervised stereo methods~\cite{Lipson2021RAFTStereoMR, li2022practical,shen2021cfnet,wang2024cost, zeng2021deep, zhang2023active} to improve model robustness against common perturbations, its application in self-supervised stereo methods remains limited. 
This is mainly because data augmentation strategies such as random \textit{color/illumination transformations} and \textit{asymmetric occlusion} may disrupt pixel-level photometric consistency assumption, leading to performance degradation.
To address this, we propose two novel self-supervised learning strategies to enhance model robustness and context awareness of \textit{illumination changes} and \textit{occlusions}. 
First, we incorporate a data augmentation branch that applies diverse transformations to the standard branch.
We then construct two self-supervisory signals by contrasting the outputs of both branches.
The feature-level stereo contrastive loss enforces consistency for learned features under \textit{illumination variations}. The image-level disparity difference loss compares outputs from the standard and augmented branches to ensure low sensitivity and context-awareness against common disturbances such as \textit{occlusions}.
The intuition is that the augmented image pairs contain the same context information as the standard image pairs, acting as hard positive samples, and thus should share the same feature representation and disparities as the standard ones.

In summary, our main contributions are fourfold as follows:
\begin{itemize}
    \item We propose a general self-supervised stereo matching framework called SMFormer, where extra priors of VFMs and data augmentation consistency can provide reliable guidance to mitigate the limitation of the photometric consistency in challenging regions.
    \item We integrate VFM with the MLA mechanism into FPN to improve the representation ability of existing feature extractors.
    \item We develop two self-supervised losses to strengthen the model to be more robust and context-aware against common real-world disturbances.
    \item Results on multiple benchmarks demonstrate that {SMFormer} achieves highly competitive performance, even comparable with SOTA supervised methods.
Remarkably, in the challenging Booster benchmark~\cite{ramirez2022open}, {SMFormer} even outperforms recent SOTA supervised methods, such as CFNet.
\end{itemize}




\section{Related Work}
\label{sec2}
\subsection{Learning-based Stereo Matching}
Learning-based methods have shown significant improvements in stereo matching. Recently, RAFT-stereo~\cite{Lipson2021RAFTStereoMR}, IGEV-Stereo~\cite{xu2023iterative}, and Selective-IGEV~\cite{wang2024selective}
develop multi-level GRUs to achieve impressive performance iteratively. 
\textcolor{black}{With the emergence of Visual Foundation Models (VFMs), transferring rich priors has further boosted downstream tasks, including stereo matching. Several methods~\cite{jiang2025defom,zhanglearning2024,wen2025stereo,zhou2025all,wang2025learning} have begun to incorporate pre-trained VFMs into existing stereo backbones. For example, DEFOM-Stereo~\cite{jiang2025defom} leverages strong monocular depth priors from a depth foundation model to achieve competitive results. Among these, the most relevant are SMoE-Stereo and FoundationStereo. SMoE-Stereo adapts VFMs for robust in-the-wild stereo matching via a selective MoE mechanism that dynamically activates scene-appropriate experts in a parameter-efficient manner.
FoundationStereo~\cite{wen2025stereo}, in contrast, pre-trains on one million synthetic stereo pairs to obtain a stereo foundation model that enables zero-shot generalization in the wild.
Compared with these methods, SMFormer also leverages robust priors from VFMs, yet it diverges fundamentally in both motivation and design.
In terms of motivation, AIO-Stereo and FoundationStereo depend heavily on dense ground-truth annotations to improve performance and generalization, whereas SMFormer is designed to remain effective in real-world scenarios without reliance on labeled data.
In terms of design, AIO-Stereo integrates multiple VFMs to pursue maximal accuracy, and FoundationStereo constructs a large, general-purpose stereo foundation model. In contrast, SMFormer introduces a lightweight multi-layer attention mechanism to efficiently transfer priors from VFMs while emphasizing designing effective self-supervised loss functions.}

\subsection{Self-supervised Stereo Matching}
\textcolor{black}{Since deep stereo matching methods require dense ground-truth labels, self-supervised stereo matching has emerged as a promising paradigm to alleviate the reliance on costly disparity annotations.
Zhong \textit{et al.}~\cite{zhong2017ssl} first proposed a loop photometric consistency loss.
Wang \textit{et al.}~\cite{Unos2019} and Liu \textit{et al.}~\cite{liu2020flow2stereo} jointly learned optical flow and stereo matching from video stereo images. 
Li \textit{et al.}~\cite{li2021unsupervised} and Chen \textit{et al.}~\cite{chen2021revealing} used occlusion cues for stereo matching. 
Wang \textit{et al.}~\cite{wang2020parallax} proposed a parallax attention mechanism to capture the stereo correspondence without limiting disparity variations.
Su \textit{et al.}~\cite{su2022chitransformer} used monocular cues and vision transformer~\cite{dosovitskiy2020image} (ViT) with cross-attention to enhance binocular depth estimation.
Tosi \textit{et al.}~\cite{tosi2023nerf} utilized NeRF to generate thousands of stereo pairs for training.
However, these self-supervised approaches primarily rely on the assumption of photometric consistency, which often leads to ambiguous supervision in challenging scenarios. To address this limitation, recent studies~\cite{xu2021self,chang2022rc,wang2025dualnet,wang2025rose} have introduced more reliable supervisory signals into multi-view stereo. The most related work is JDACS~\cite{xu2021self}, which incorporates a data-augmentation consistency loss to mitigate color variations. Unlike JDACS, we tackle the more complex challenges of photometric consistency, which are influenced not only by color variations but also by reflections, occlusions, and textureless surfaces.}

\subsection{Feature Learning in Stereo Matching}
Feature representation learning is critical in finding accurate pixel-wise correspondence between rectified stereo pairs.
As a common solution for feature extraction, a Feature Pyramid Network (FPN) with different dilation settings is proposed~\cite{chang2018pyramid}. Then, a modified feature extraction mechanism is adopted in~\cite{shen2021cfnet,shen2022pcw,guo2022cvcnet,wang2025adstereo}.
Some work embeds deformable convolutions~\cite{xu2020aanet,zhang2021hda}, attention mechanisms~\cite{zhao2023high} into CNN-based networks to learn more reliable features.
Most recently, ViTAStereo~\cite{liu2024playing} and FormerStereo~\cite{zhanglearning2024} have introduced Vision Transformer (ViT)-based Visual Foundation Models (VFMs) as robust feature extractors for stereo matching, achieving impressive performance. These models rely on lightweight adapters to retrieve fine-grained target information from general-purpose coarse-grained tokens. However, they may struggle to capture rich target domain information, potentially leading to sub-optimal performance~\cite{goyal2023finetune}. To address this, we propose to integrate the strengths of CNNs and ViTs, as features extracted from VFMs are complementary to those obtained from Feature Pyramid Networks (FPN).

\subsection{Visual Foundation Models (VFMs)}
Visual Foundation Models (VFMs) have recently become a key solution for reducing reliance on annotated data while enhancing model robustness. Leveraging extensive training datasets~\cite{kirillov2023segment, yang2024depth_v2, weinzaepfel2023croco} and advanced self-supervised learning techniques~\cite{oquab2024dinov2}, VFMs are revolutionizing the field.
Among these, DUST3R~\cite{dust3r_cvpr24} stands out as the first unified 3D vision pipeline. The SegmentAnything Model (SAM)~\cite{kirillov2023segment}, which inspired DepthAnything (DAM) and  DepthAnythingV2 (DAMV2)~\cite{depthanything, yang2024depth_v2}, exemplifies a versatile segmentation model with promising zero-shot transfer capabilities across downstream tasks. Concurrent works like EVA2~\cite{fang2023eva} and DINOV2~\cite{oquab2024dinov2} also highlight the use of VFMs for diverse visual applications.
Despite the appeal of VFMs in various fields, unlocking their potential in stereo matching remains challenging. In this work, we aim to leverage the informative priors of VFMs to enhance feature learning.

\subsection{Contrastive Learning}
Contrastive learning~\cite{he2020momentum, chen2021exploring, wang2022contrastive} aims to encourage models to maintain invariance across various data augmentations of a single instance. 
By bringing positive sample pairs closer and pushing negative sample pairs further apart, it significantly reduces the performance gap between self-supervised and supervised models.
Recent advancements, such as MoCo and its variants~\cite{he2020momentum, chen2020improved}, conceptualize contrastive learning as a dictionary look-up process using a momentum-updated queue encoder. Several approaches~\cite{xie2021propagate, zhang2022revisiting} have extended this framework to construct hard positive samples, thereby enhancing pixel-level feature learning.
Inspired by these, our method introduces two additional supervisory signals at both intermediate and output stages for self-supervised stereo matching by employing the priors of data augmentation consistency.

\section{Methodology}
\label{sec3}
\begin{figure*}[t]
\includegraphics[width=1.0\linewidth]{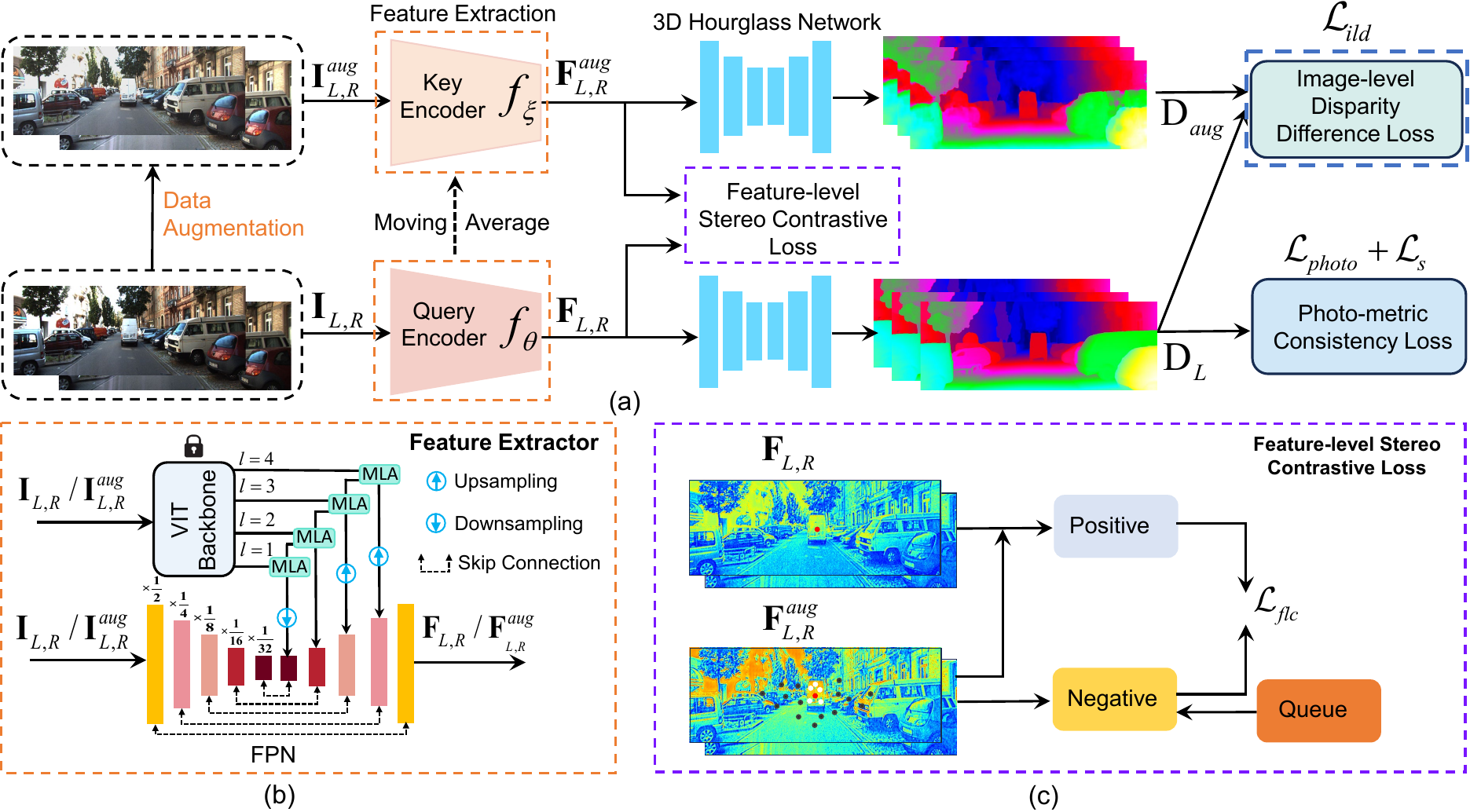}
\vspace{-0.6cm}
    \caption{An overview of {SMFormer}. During training, the framework adopts a data augmentation branch with an augmented pair 
    (the upper part of subfigure (a)) and a standard branch with a standard pair (the lower part of subfigure (a)). Only the standard branch is available for inference.
    Both branches share weights in the cost aggregation module except for the feature extractors.
    For the feature-level stereo contrastive loss, the key encoder for the augmented image pair is a moving average of the query encoder to mitigate negative sample effects.  The details regarding the feature extractor are in subfigure (b). Color-dashed boxes are our main contributions.}
      \vspace{-0.3cm}
    \label{sec3:overview}
\end{figure*}


In this section, the main contributions of SMFormer are elaborated. 
We first depict the self-supervised framework (Sec.~\ref{sec3:preliminary}). Then, we describe the proposed feature extractor (Sec.~\ref{sec3:2}). The proposed feature-level stereo contrastive loss, and the proposed image-level disparity difference loss (Sec.~\ref{sec3:3} \& Sec.~\ref{sec3:4}).
Finally, we introduce the overall loss function during training (Sec.~\ref{sec3:5}).
Our SMFormer comprises a standard branch and a data augmentation branch, adopting CFNet~\cite{shen2021cfnet} as the backbone.
A sketch of the pipeline is shown in Fig.~\ref{sec3:overview} (a).
Note that SMFormer is a general framework suitable for arbitrary
learning-based stereo matching. 
We only integrate the VFM (i.e., SAM~\cite{kirillov2023segment}) with the MLA mechanism into FPN to replace its feature extractor while the remaining structures are unchanged.

\subsection{Self-supervised Stereo Matching}
\label{sec3:preliminary}
\noindent\textbf{Network Details.}
Given a rectified stereo image pair ${\bf{I}}_{L,R}$, our goal is to train the framework to estimate the corresponding disparity map $\textbf{D}_{L}$ without the Ground Truth label. 
Specifically, the framework consists of four steps: feature extraction, cost volume construction, cost volume aggregation, and disparity regression.
During feature extraction, the network first uses the proposed feature extractor to extract multi-scale features ${\textbf{F}}_{L}, {\textbf{F}}_{R}\in{\mathbb{R}^{\frac{H}{s}\times \frac{W}{s}}}$ ($s\in\left\{2, 4, 8\right\}$) from stereo image pairs.
The multi-scale left and right features are next formed into a cascade cost volume at three stages, with incremental image resolutions (1/8, 1/4, and 1/2 of the original resolution), following the approach in~\cite{shen2021cfnet}:
\begin{align}
& \textbf{C}_{cat}^{i}(f,d^{i},x,y) = \textbf{F}_{L}^{i}(x,y) \parallel  \textbf{F}_{R}^{i}(x-d^{i},y) \notag\\
& \textbf{C}_{gwc}^{i}(g,d^{i},x,y) = \frac{1}{N_{c}^{i}/N_{g}} \left\langle  \textbf{F}_{L}^{i}(x,y), \textbf{F}_{R}^{i}(x-d^{i},y) \right\rangle \\
& \textbf{C}_{}^{i} = \textbf{C}_{cat}^{i} \parallel \textbf{C}_{gwc}^{i} \notag
\end{align}
where $d^{i}$ is the disparity hypothesis, $||$ denotes the vector concatenation operation. $N_{c}$ indicates the number of channels in the extracted features. $N_{g}$is the number of channel groups. $<,>$ represents the inner product. $\textbf{F}^i$ refers to the extracted feature at (stage) $i$ and $i = 0$ represents the original resolution.

For each raw cost volume $\textbf{C}$, we use a regular 3D hourglass network~\cite{shen2021cfnet} and a softmax operation to produce a probability volume $\textbf{P}_{v}$.
Finally, the disparity map $\textbf{D}_{L}$ is obtained by a weighted sum:
\begin{equation}
\textbf{D}_{L}=\sum_{d_{min}}^{d_{max}}d\times \textbf{P}_{v}(d).
\end{equation}
In a coarse-to-fine fashion, the coarse disparity map $\textbf{D}_{L}$ is used to generate the disparity hypothesis $d^{i}$ of the next stage~\cite{shen2021cfnet, shen2022pcw}.
With features ${\textbf{F}}_{L,R}\in{\mathbb{R}^{\frac{H}{s}\times \frac{W}{s}}}$ at the larger resolution, finer
disparity maps $\textbf{D}_{L}$ will be estimated iteratively.

\begin{figure}[t]
    \includegraphics[width=1.0\linewidth]{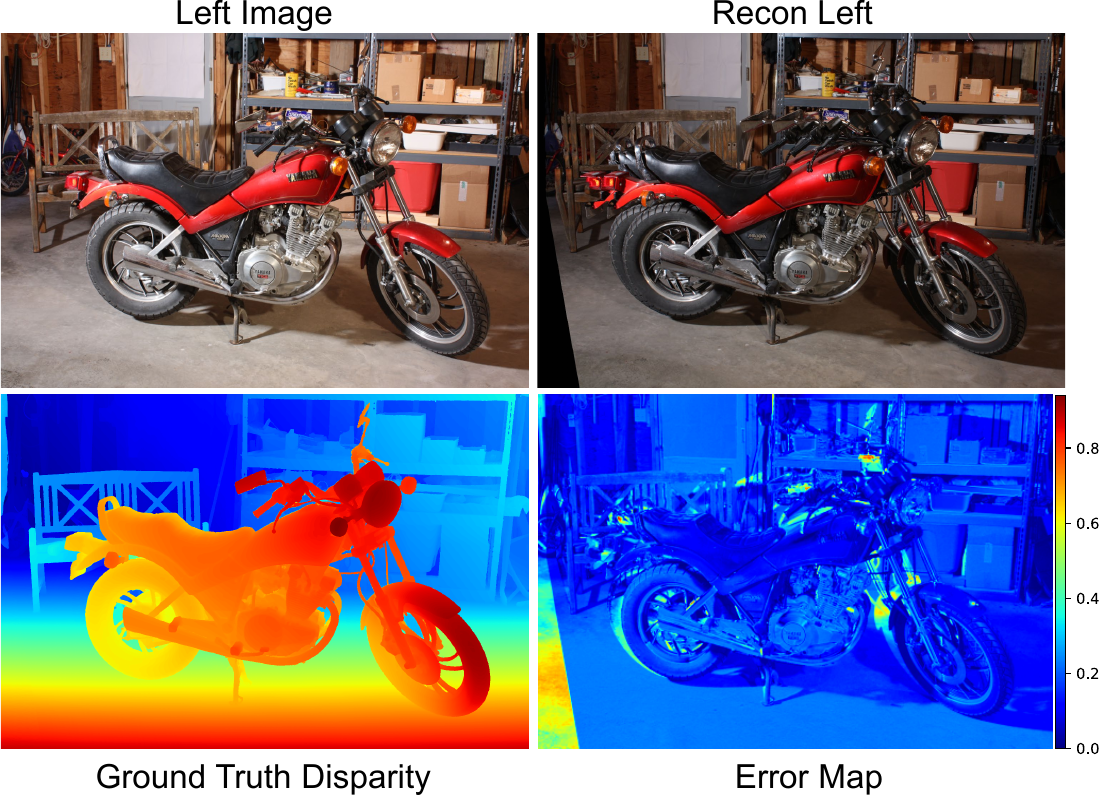}
    \vspace{-0.5cm}
    \caption{The visualization of the reconstructed left and the left image on the MotorcycleE image pair. For the Ground Truth disparity with occlusions, we use the predicted disparity values to fill in the gaps. The error map is calculated from the absolute difference between these two images.}
    \vspace{-0.1cm}
    \label{sec3:photo_metric}
\end{figure}

\noindent\textbf{Vanilla Self-supervised Loss.}
In self-supervised stereo matching, we utilize the input itself to supervise our model instead of relying on expensive ground truth labels. With a precise dense disparity map, we can accurately reconstruct the left image by warping the right view. Thus, the left image ${\bf{I}}_{L}$ can be reconstructed from the right image ${\bf{I}}_{R}$ as follows:
\begin{equation}
     {\bf\hat{I}}_{L}(i,j) = {\bf{I}}_{R}(i+{\bf{D}}_{L}(i,j),j), 
     \label{warp_loss}
\end{equation}
where (\textit{i, j}) represents the pixel coordinates and ${\bf\hat{I}}_{L}$ is the reconstructed left image. 
The hypothesis of photometric consistency aims to maximize the similarity between the left image~${\bf{I}}_{L}$ and the reconstructed left image ${\bf\hat{I}}_{L}$ after being warped from the right perspective. 
Accordingly, the photometric consistency loss can be formulated as:
\begin{equation}
    \begin{split}
    \mathcal{L}_{photo} = \frac{1}{N}\sum_{i, j} &(\alpha\frac{1-\mathcal{S}({\bf{I}}_{L}(i,j),{\bf\hat{I}}_{L}(i,j))}{2} + \\
    &(1-\alpha)||{\bf{I}}_{L}(i,j)-{\bf\hat{I}}_{L}(i,j)||),
    \end{split}
     \label{photo_loss}
\end{equation}
where $N$ is the number of pixels, $\mathcal{S}$ is an SSIM function~\cite{wang2004image}, and hyper-parameter $\alpha$ is empirically set to 0.85.
In real-world scenarios, stereo matching images are captured by a pair of cameras, leading to inevitable discrepancies in visibility, brightness, and color. These scene asymmetries reduce the effectiveness of photometric consistency loss. 
As illustrated in Fig.~\ref{sec3:photo_metric}, even when using processed ground truth disparity (with occlusion filled) to reconstruct the left image from the low-light right view, the error map remains significant. Such asymmetric noise misguides the network, leading to undesirable results during training. 
Therefore, in self-supervised stereo matching, it is crucial to incorporate valuable priors to address disparity estimation in challenging regions, rather than relying solely on the photometric consistency loss.

Besides, following \cite{wang2020parallax,li2023sense}, we use an edge-aware smoothness loss $\mathcal{L}_{s}$ to encourage the local smoothness of disparity:
\begin{equation}
\begin{split}
\mathcal{L}_{s}=\frac{1}{N}\sum_{i,j}&(||\nabla_{x}\textbf{D}_{L}(i,j)||_{1}\exp^{(-||\nabla_{x}\textbf{I}_{L}(i,j)||_{1})} + \\
&||\nabla_{y}\textbf{D}_{L}(i,j)||_{1}\exp^{(-||\nabla_{y}\textbf{I}_{L}(i,j)||_{1})}),
\end{split}
\end{equation}
where $\nabla_{x}$ and $\nabla_{y}$ are gradients in the $\mathrm{x}$ and $\mathrm{}{y}$ axes.
By optimizing the smoothness loss, the disparity map can be effectively aligned with the edge structure of the left image, while maintaining overall smoothness in the disparity map.

\subsection{Feature Extraction}
\label{sec3:2}
Current CNN-based feature extractors struggle to learn discriminative features from \emph{reflective} and \emph{texture-less} regions due to limited receptive fields and weak supervision. To address this issue, we integrate the pre-trained Visual Foundation Model (VFM), such as SAM~\cite{kirillov2023segment}, with a Feature Pyramid Network (FPN) in SMFormer. The VFM provides robust global feature understanding, while the FPN focuses on detailed features in the target domain.

\begin{figure*}[t]
\begin{minipage}{\textwidth}
    \begin{minipage}[h]{0.46\textwidth}
    \flushleft
\includegraphics[width=1\linewidth]{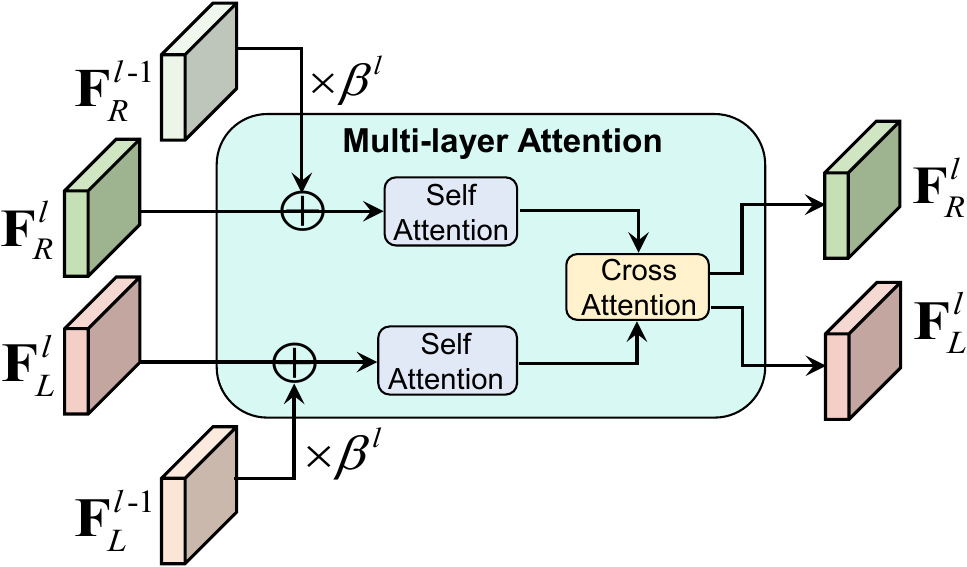}
    \flushleft
    \vspace{-0.03cm}
    \caption{Self and cross-view attentions are used to learn left and right image features, respectively. Adaptive layer modulation aims to learn the importance of various transformation layers.
    The first MLA module does not include layer modulation.}
    \label{sec3:msa}
     \end{minipage}
\begin{minipage}[h]{0.52\textwidth}
    \flushright
\includegraphics[width=1\linewidth]{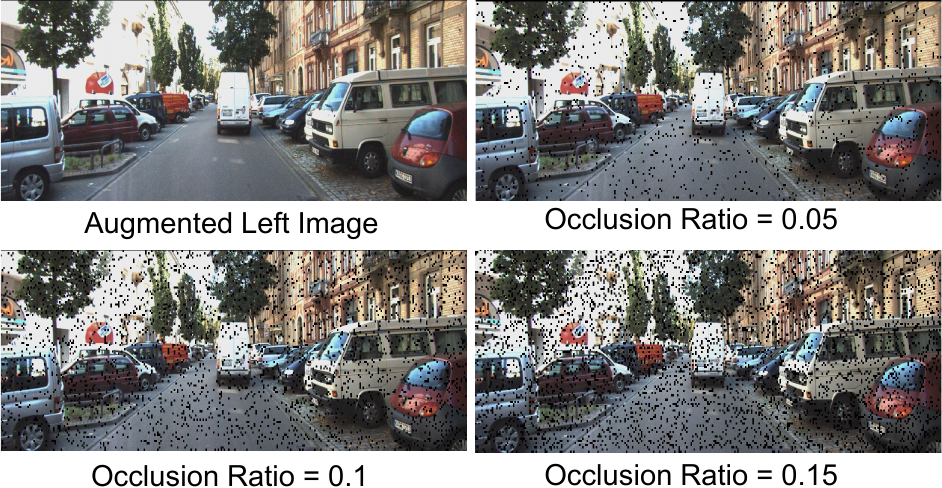}
\vspace{-0.48cm}
    \caption{Illustration of mask sampling strategy. It determines the difficulty of the reconstruction task and affects the reconstruction quality. More masked pixels mean more difficult failure cases of photometric consistency in stereo matching.}
\label{sec3:image_contrastive}
     \end{minipage}
\end{minipage}
\vspace{-0.2cm}
\end{figure*}

Given a rectified stereo image pair ${\bf{I}}_{L,R}$ or ${\bf{I}}^{aug}_{L,R}$, {SMFormer} uses the proposed feature extractor to extract features, resulting in outputs ${\bf{F}}_{L,R}$ and ${\bf{F}}^{aug}_{L,R}$, as shown in Fig.~\ref{sec3:overview} (b).
Specifically, before processing the stereo image pair with the pre-trained VFM, we resize its position embedding using bicubic interpolation to fit different image sizes~\cite{dosovitskiy2020image}. 
Following~\cite{zhanglearning2024}, features from transformation blocks at varying depths (e.g., 2nd, 5th, 8th, 11th) are then collected as the outputs at different layers $l$.
To enrich VFM features at various depths with cross-view and cross-layer contextual information, we propose a Multi-layer Attention (MLA) mechanism enhanced with adaptive layer modulation, as depicted in Fig.~\ref{sec3:overview} (b).
As detailed in Fig.~\ref{sec3:msa}, for VFM features at the $l$-th layer, we incorporate features from the previous layer $\textit{l}-1$, which are multiplied by the adaptive layer modulation coefficient $\beta^{l}$. These VFM features are then encoded using self-attention and cross-attention mechanisms before being input to the next MLA block.
The aggregated features at low resolution (1/16 of the original image resolution) undergo bilinear interpolation to yield feature outputs at four diverse scales (1/32, 1/16, 1/8, and 1/4). Subsequently, these multi-scale VFM features are concatenated with those from an FPN encoder, and a standard FPN decoder produces the final multi-scale outputs ${\textbf{F}}_{L},{\textbf{F}}_{R}\in{\mathbb{R}^{\frac{H}{s}\times \frac{W}{s}}}$ ($s\in\left\{2, 4, 8\right\}$). 
These features, which contain rich content information from both ViT and FPN, are used to formulate an informative and discriminative cascade cost volume, as clarified in Sec.~\ref{sec3:preliminary}.

\subsection{Feature-level Stereo Contrastive Loss}
\label{sec3:3}
The learned stereo features from the proposed feature extractor are used to construct the cost volume, which is a crucial internal representation within a deep stereo network. 
To improve the network's robustness against changes in illumination, we enforce a consistency constraint on the stereo features.
Inspired by contrastive learning for discriminative feature representation, we develop a novel contrastive learning mechanism for stereo features.
Specifically, we apply a pixel-wise contrastive loss to stereo feature pairs and use a dictionary queue with a momentum-updated key encoder $f_{\xi}$ to store diverse negative samples across the datasets.

\noindent\textbf{Inputs.}
As shown in Fig.~\ref{sec3:overview} (a),  
an input image pair is first pre-processed by two different branches $x_{1}$ $\sim$ $\mathcal{T}_{1}(x)$ (the standard branch) and $x_{2}$ $\sim$ $\mathcal{T}_{2}(x)$ (the data augmentation branch), which are then fed into the query encoder $f_{\theta}$
and the key encoder $f_{\xi}$ to obtain ${\bf{F}}_{L,R}$ and ${\bf{F}}^{aug}_{L,R}$,  where $\theta$ represents the learnable parameters and $\xi$ is the exponential moving average of $\theta$.
For $\mathcal{T}_{2}(x)$, we adopt a data augmentation strategy that performs diverse transformations.
\textcolor{black}{These transformations include variations in brightness, gamma, and contrast to simulate asymmetric illumination changes, Glass blur to model reflective regions, Gaussian blur to approximate low-texture areas, and occlusion masks to represent occluded regions.} 

 \begin{figure}[t]
    \includegraphics[width=1.0\linewidth]{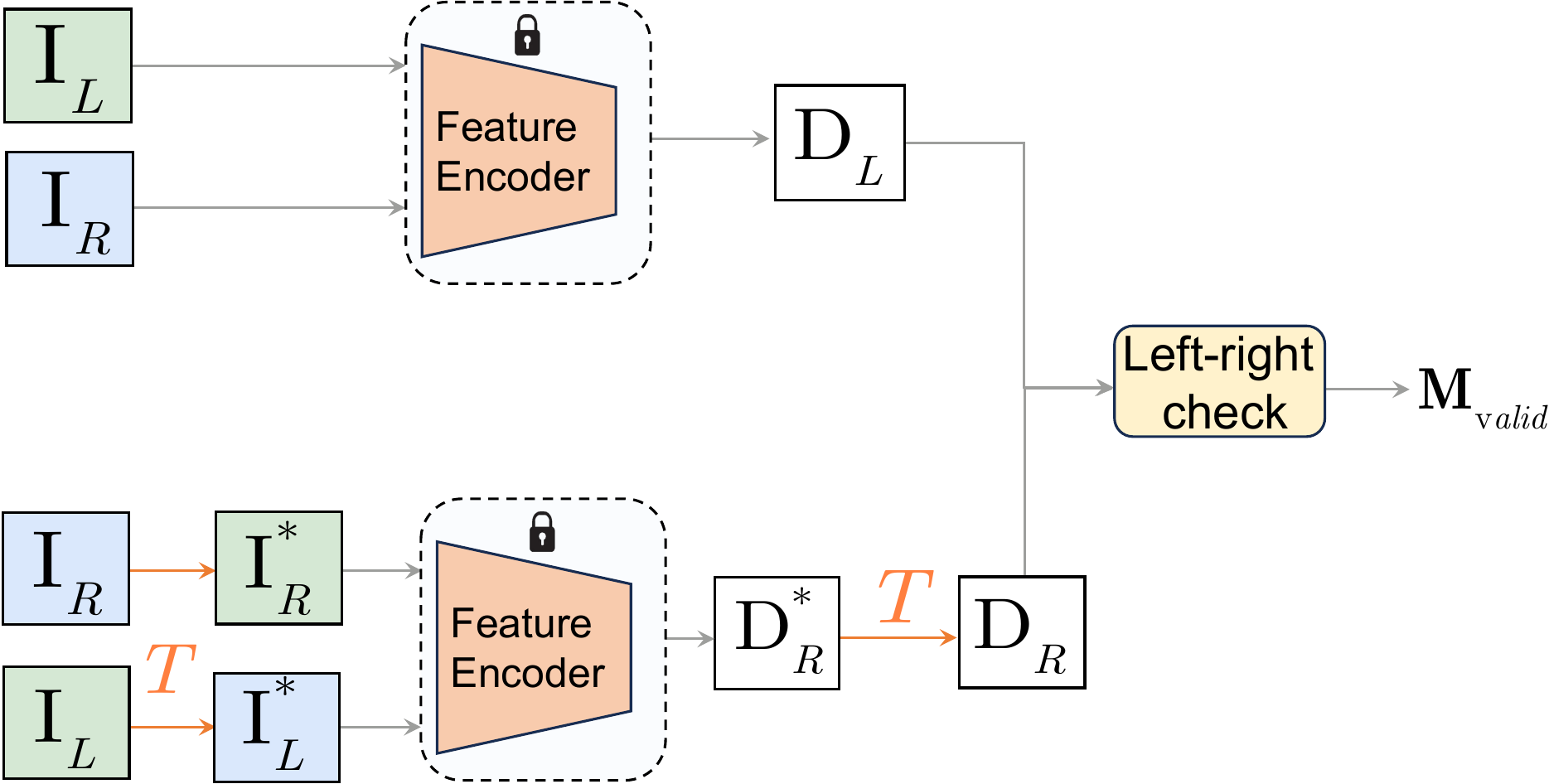}
    \caption{A pipeline of valid checks. $\textit{\textcolor{orange}{T}}$ denotes the horizontal flip operation.}
    \label{fig:valid_mask}
    \vspace{-0.2cm}
\end{figure}

\noindent\textbf{Positive and Negative Pairs.}
Contrastive learning aims to construct positive and negative samples, in which the representations in the positive samples stay close to each other while the negative ones are far apart.
We first define the query pair features $\mathcal{Q}^{L, R}_{\textit{\textbf{p}}}$ in the standard branch as the anchor points (red points in Fig.~\ref{sec3:overview} (c)), where $\textit{\textbf{p}}$ denotes pixel position.
We consider the pixel vectors in the data augmentation branch as positive pairs $\mathcal{K}^{+}$  if their pixel coordinates are close to the pixels in the standard branch (white points in Fig.~\ref{sec3:overview} (c)),
\textit{i.e.,} the query pair features $\mathcal{Q}^{L, R}_{\textit{\textbf{p}}}$ in the standard branch are paired with the key pair features $\mathcal{K}^{L, R}_{\textit{\textbf{p}}+\Delta \textit{\textbf{p}}}$ in the data augmentation branch, where $\Delta\textit{\textbf{p}}$ $\in [-1,1]$ is a 2D vector and indicates the random location offsets.
Meanwhile, negative pairs $\mathcal{K}^{-}$ (black points in Fig.~\ref{sec3:overview} (c)) are randomly sampled from $N$ non-matching points with large offsets from the key pair features $\mathcal{K}^{L, R}_{\textit{\textbf{p}}}$ within a window size $50\times50$, forming $N$ negative pairs.

\noindent\textbf{Momentum Encoder.}
The choice of negative samples significantly impacts learned representations~\cite{xie2021propagate, zhang2022revisiting}. 
Typically, it is more intuitive and effective to use pixels from different image pairs as negative samples, as this approach aligns well with the concept of a dictionary queue. To implement this, we adopt the method described in~\cite{he2020momentum}, which involves maintaining a dynamic dictionary queue that acts as a repository for storing the representations of the entire dataset. During each mini-batch, the dictionary randomly selects a fixed number of samples, denoted as 
$K$, from this memory bank. Importantly, this sampling process does not involve back-propagation, ensuring computational efficiency.
However, this method will make the key representations inconsistent when sampling~\cite{he2020momentum}.
To address this, we replace weight-sharing feature extractors with an asymmetric pair: a query encoder and a momentum-based moving average key encoder, which evolves gradually with the query encoder.
\begin{equation}
    \bm{\xi}_{t} = m \bm{\xi}_{t-1} + (1-m) \bm{\theta}_{t},
    \label{sec3:eq1}
\end{equation}
where $t$ is the number of iterations and $m\in[0,1)$ is a momentum coefficient commonly set to 0.999. 
We use back-propagation to update the parameter $\bm{\theta}$ and use Eq.~\ref{sec3:eq1} to update~$\bm{\xi}$. 
According to the different update schemes, the query and key will ultimately be encoded by separate encoders.

\noindent \textbf{Pixel-wise Contrastive Loss.}
Using the provided positive and negative pairs, we calculate the similarity of feature pairs through the dot product and employ a pixel-wise InfoNCE loss~\cite{he2020momentum}:
\begin{equation}
\mathcal{L}_{flc} = - \rm{log} \frac{exp(\mathcal{Q}^{L, R}_{\textit{\textbf{p}}}\cdot\mathcal{K}^{+}/\tau)}{\sum_{\mathcal{K}} exp(\mathcal{Q}^{L, R}_{\textit{\textbf{p}}}\cdot \mathcal{K}^{-})/\tau},
\label{sec3:eq2}
\end{equation}
where $\tau$ is a temperature hyper-parameter. The sum is over one positive and $K$ negative samples. Following~\cite{zhang2022revisiting}, we empirically set $N$ = 60, $K$ = 6000, and $\tau$ = 0.07.

\subsection{Image-level Disparity Difference Loss}
\label{sec3:4}

The feature-level stereo contrastive loss in Sec.~\ref{sec3:2} helps maintain feature consistency regardless of illumination variations.
However, the existing supervisory functions fail to offer valid supervisory signals in ill-posed regions, e.g., \textit{occluded} regions. 
Actually, we can treat these failures in photometric consistency as hard positive samples and better utilize them in a similar contrastive learning manner to improve the model's context-awareness. 
To construct these failures,
inspired by Masked Image Modeling~\cite{he2022masked, rao2023masked}, we utilize a uniform distribution to apply random masking to the pixels in the augmented left image, denoted as augmented left image ${\bf I}^{aug}_{L}$. 
The occlusion mask has a fixed small size and a gradually increased occlusion ratio $\alpha$ (0 to 0.15), as shown in Fig.~\ref{sec3:image_contrastive}.
The hard positive sample pair is then fed into the augmented branch to infer the augmented disparity map $\mathbf{D}_{aug}$. The prediction from the standard branch is denoted as ${\bf D}_{L}$. 
To enforce output consistency between ${\bf D}_{L}$ and ${\bf D}_{aug}$, the image-level disparity difference loss uses $Smooth_{L_1}$ loss term to minimize the differences between them  due to its robustness against outliers~\cite{ren2016faster}:
\begin{equation}
    \mathcal{L}_{ild} = \rm{Smooth_{L_1}}\left({\bf D}_{\textit{aug}}\odot \textbf{M}_{\textit{v}}, \; {\bf D}_{\textit{L}} \odot \textbf{M}_{\textit{v}}\right),
    \label{sec3:eq3}
\end{equation}
where the binary valid mask $\textbf{M}_{v}$ is used to filter the unreliable point.
To ensure training stability, we employ curriculum learning to increase the occlusion rate $\alpha$ gradually, which grows from 0 to 0.15 in our implementation.

\noindent \textbf{Valid Mask.}
Given that the disparity map $\textbf{D}_{L}$ may contain outliers, the model might focus on potentially erroneous regions. 
To identify these poorly estimated regions, we use a classic left-right consistency check to filter outliers~\cite{hirschmuller2007stereo}, as illustrated in Fig.~\ref{fig:valid_mask}. We apply a horizontal flip transformation to the original image pairs and input them into the model to estimate the right disparity map, ${\bf D}_{R}$.
For any point $\textbf{\textit{p}}(i,\ j)$ in the left image, if its disparity in ${\bf D}_{L}$ is $d_{l}$, the corresponding point $\textbf{\textit{p}}(i-d_{l},\ j)$ in ${\bf D}_{R}$ has disparity $d_{r}$. The warp error at point $\textbf{\textit{p}}$ is $e_{warp}=||d_{l}-d_{r}||$. Thus, $\textbf{M}_{v}$ can be represented as
\begin{equation}
    {\bf M}_{v} = \begin{cases}
	1, &\; 
    e_{warp} \leq \tau_{warp} \\
	0, &\; otherwise\\
    \end{cases},
    \label{sec3:lrc}
\end{equation}
where $\tau_{warp}$ represents threshold values and we set $\tau_{warp}$ to 3 pixel in our settings.

\subsection{The Overall Training Objectives}
\label{sec3:5}

To summarize, the final objectives are as follows:
\begin{equation}
    \mathcal{L} = \lambda_{1}\mathcal{L}_{photo}  + \lambda_{2}\mathcal{L}_{s} +
    \lambda_{3}\mathcal{L}_{flc} + \lambda_{4}\mathcal{L}_{ild},
\end{equation}
where $\mathcal{L}_{photo}$ is the photometric loss and $\mathcal{L}_{s}$ is the disparity smooth loss~\cite{wang2020parallax}. $\lambda_{i}$ is a hyper-parameter set where $\lambda_{1}$ = $\lambda_{3}$ = $\lambda_{4}$ = 1 and $\lambda_{2}$ = 10. 
Please note that only adopting $\mathcal{L}_{flc}$ and $\mathcal{L}_{ild}$ is not feasible, as it may lead to training collapses with the learned features and disparity maps converging to constant values (typically 0)~\cite{ding2022kd}.

\section{Experiments}
\subsection{Experimental Settings}
\subsubsection{Datasets and Metrics}
\textbf{SceneFlow~\cite{dosovitskiy2015flownet}} is a synthetic dataset comprising 35454 training pairs and 4,370 evaluation pairs of stereo images, all with a resolution of  $540\times960$.
We randomly cropped patches of size $320 \times 640$ for training.

\noindent\textbf{KITTI 2012 \& 2015~\cite{geiger2012we, menze2015object}} collects outdoor driving scenes with sparse ground-truth disparities. For KITTI 2012, it contains 194 training samples and 195 testing samples with a resolution of $370\times1226$. KITTI 2015 contains 200 training samples and 200 testing samples with a resolution of $375\times1242$.
Following previous works~\cite{wang2020parallax,liu2020flow2stereo}, the mixed KITTI 2012 and 2015 training sets are used for training (394 image pairs). 
We randomly cropped patches of size $320 \times 832$ for training.
The maximum disparity range is set to 192.

\begin{table*}[t]
\caption{Comparative results achieved on the KITTI 2012 \& 2015 benchmarks. The latest state-of-the-art self-supervised stereo matching method is ChiT but using KITTI Eigen Splits (22600 image pairs), denoted as $\ast$.
“-” indicates that results are not available.}
\vspace{-0.4cm}
    \begin{center}
    \scriptsize
	\setlength{\tabcolsep}{2.5mm}{
		\begin{tabular}{c|c|cccc|ccc|c}
		\toprule
		& & \multicolumn{4}{c|} {\textbf{KIT 2012}}  & \multicolumn{3}{c|}{\textbf{KIT 2015}}  \\
		\multirow{1}{*}[5pt]{Method} & \multirow{1}{*}[5pt]{\textbf{Supervised.}} & \textbf{Out-Noc (\%)} & \textbf{Out-All (\%)} & \textbf{Avg-Noc (px)} & \textbf{Avg-All (px)} & \textbf{D1-bg (\%)} & \textbf{D1-fg (\%)} & \textbf{D1-All (\%)} & \textbf{Time (s)}\\
		\midrule
     DispNet~\cite{dosovitskiy2015flownet} & \Checkmark & 4.11 & 4.65 & 0.9 & 1.0 & 4.32 & 4.41 & 4.34 & 0.06 \\
    GCNet~\cite{kendall2017end} & \Checkmark & 1.77 & 2.30 & 0.6 & 0.7 & 2.21 & 6.16 & 2.87 & 0.9 \\
    MABNet~\cite{xing2020mabnet} & \Checkmark & 2.71 & 3.31 & 0.7 & 0.8 & 3.04 & 8.07 & 3.88 & 0.11 \\
    SGM-Net~\cite{seki2017sgm} & \Checkmark & 2.29 & 3.50 & 0.7 & 0.9 & 2.66 & 8.64 & 3.66 & 67 \\
	\midrule
    OASM~\cite{li2018occlusion} & \XSolidBrush & 6.39 & 8.60 & 1.3 & 2.0 & 6.89 & 19.42 & 8.98 & 0.73 \\
PASMnet\_192~\cite{wang2020parallax} & \XSolidBrush & 7.14 & 8.57 & 1.3 & 1.5 & 5.41 & 16.36 & 7.23 & 0.5 \\
Flow2Stereo~\cite{liu2020flow2stereo} & \XSolidBrush & 4.58 & 5.11 & 1.0 & 1.1 & 5.01 & 14.62 & 6.61 & 0.05 \\
DispSegNet~\cite{zhang2019dispsegnet} & \XSolidBrush & 4.68 & 5.66 & 0.9 & 1.0 & 4.20 & 16.67 & 6.33 & 0.9 \\
ChiT$^{\ast}$~\cite{su2022chitransformer} & \XSolidBrush & - & -& - & - & \textbf{2.50} & \textbf{5.49} & \textbf{3.03} & 0.5 \\
    \textbf{SMFormer (Ours)} & \XSolidBrush & \textbf{2.97} & \textbf{3.57} & \textbf{0.7} & \textbf{0.8} & \textbf{\textcolor{black}{2.79}} &  \textbf{\textcolor{black}{8.17}} &  \textbf{\textcolor{black}{3.68}} & 0.3 \\
\bottomrule
\end{tabular}}
\end{center}
\vspace{-0.4cm}
\label{sec4:table_kitti}
\end{table*}

\begin{figure*}[t]
\includegraphics[width=1.0\linewidth]{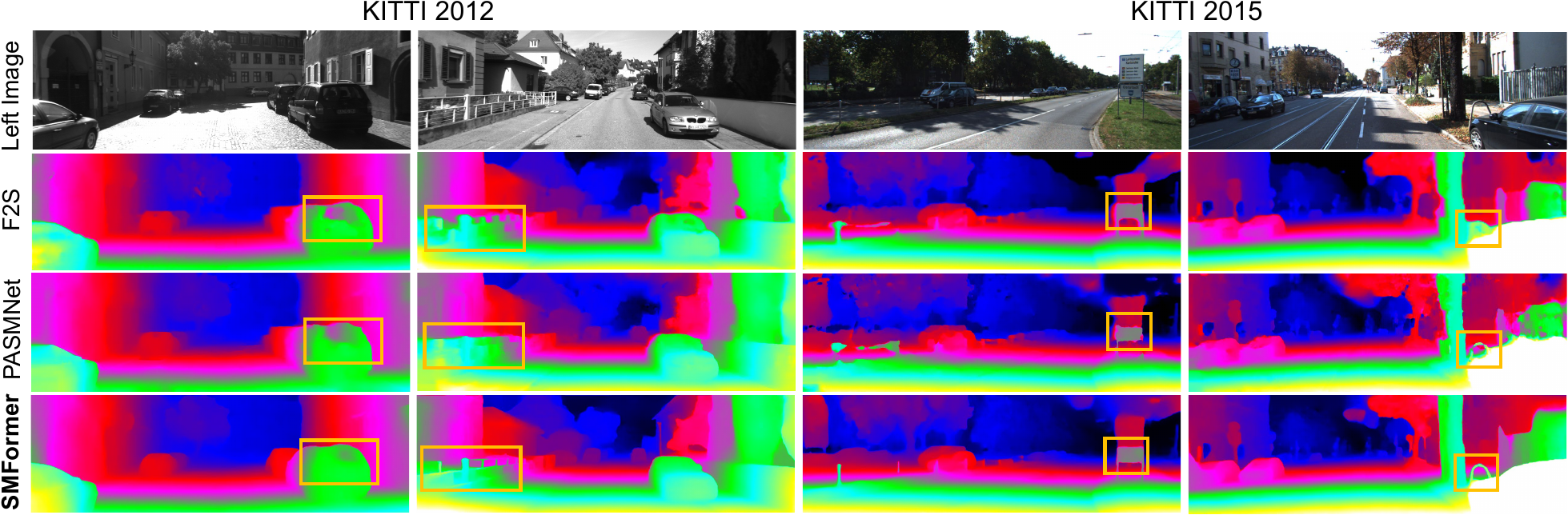}
    \vspace{-0.5cm}
    \caption{Visualization results achieved by our method and other self-supervised stereo matching methods on KITTI benchmarks. 
    Our \textbf{SMFormer} performs well in reflective and detailed regions, highlighted by \textcolor{orange}{orange} boxes.  Best zoomed in.}
    \vspace{-0.3cm}
    \label{sec4:vis_kit}
\end{figure*}

\begin{table}[t]
\centering
\scriptsize
\caption{Domain generalization evaluation on target training sets. $^\ddag$ indicates the use of pre-trained VFM.  The best results in \textbf{bold} and the sub-optimal best results in \textbf{\textcolor{black}{blue}}).}
\vspace{-0.2cm}
\setlength{\tabcolsep}{.1mm}{
    \begin{tabular}{l|cccc}
    \toprule
    \multirow{2}{*}[-1pt]{\; Method} & {\textbf{KIT 2012}} & {\textbf{KIT 2015}}  & {\textbf{Middle}} & \textbf{ETH3D}\\
    & \textbf{Bad 3.0 (All)} & \textbf{Bad 3.0 (All)}  & \textbf{Bad 2.0 (Noc)} &  \textbf{Bad 1.0 (Noc)} \\
    \midrule
    CFNet~\cite{shen2021cfnet} & 4.7 & 5.8 & 15.4 & 5.8 \\
    PCWNet~\cite{shen2022pcw} & \textbf{\textcolor{black}{4.2}} & 5.6 & 15.8 & ${5.2}$ \\
    RAFT-Stereo~\cite{Lipson2021RAFTStereoMR} & 5.1 & 5.7 & 12.6 & 3.3\\
    IGEVStereo~\cite{xu2023iterative} & 5.7 & 6.0 & \textbf{7.2} & 4.1 \\
    UCFNet\_pretrain~\cite{2023uCFNet}  & 4.5 & 5.2 & 26.0 & 4.8 \\
    LoS~\cite{li2024local} & 4.4 & 5.5 & 19.6 & $\textbf{\textcolor{black}{3.1}}$ \\
    Former-PSMNet$^\ddag$ (SAM)~\cite{zhanglearning2024}& 4.3 & $\textbf{\textcolor{black}{5.0}}$ & $ {9.4}$ & 6.4 \\
    \midrule
    \textbf{SMFormer (Ours)} & $\textbf{4.1}$  & $\textbf{4.7}$ &  $\textbf{\textcolor{black}{8.1}}$  & \textbf{2.9}\\
    \bottomrule
    \end{tabular}}
\label{sec4:dg_comparison}
\vspace{-0.3cm}
\end{table}

\begin{table*}[t]
\centering
\scriptsize
\begin{minipage}{0.48\linewidth}
    \centering
    \caption{Zero-shot performance on DrivingStereo under different weather conditions, evaluated using official weights and the D1 metric.}
    \vspace{-0.2cm}
    \setlength{\tabcolsep}{1mm}{
    \begin{tabular}{c|c|ccccc}
    \toprule
    Method & Venue & Sunny & Cloudy & Rainy & Foggy & Avg. \\
    \midrule
    CFNet~\cite{shen2021cfnet} & CVPR'21 & 5.4 & 5.8 & 12.0 & 6.0 & 7.3 \\
    PCWNet~\cite{shen2022pcw} & ECCV'22 & 5.6 & 5.9 & 11.8 & 6.2 & 7.4 \\
    DLNR~\cite{zhao2023high} & CVPR'23 & 27.1 & 28.3 & 34.5 & 29.0 & 29.8 \\
    IGEV-Stereo~\cite{xu2023iterative} & CVPR'23 & 5.3 & 6.3 & 21.6 & 8.0 & 10.3 \\
    Selective-IGEV~\cite{wang2024selective} & CVPR'24 & 7.0 & 8.0 & 18.4 & 12.9 & 11.1 \\
    MochaStereo~\cite{chen2024mocha} & CVPR'24 & 12.8 & 27.4 & 24.6 & 22.8 & 21.9 \\
    Former-CFNet$^\ddag$~\cite{zhanglearning2024} & ECCV'24 & 3.8 & \textbf{2.7} & \textcolor{black}{\textbf{8.3}} & 5.2 & \textcolor{black}{\textbf{5.0}} \\
    DEFOMStereo$^\ddag$~\cite{jiang2025defom} & CVPR'25 & \textcolor{black}{\textbf{3.6}} & 3.8 & 13.5 & \textbf{2.9} & 6.0 \\
    \midrule
    \textbf{SMFormer} & - & \textbf{3.2} & \textcolor{black}{\textbf{3.0}} & \textbf{6.2} & \textcolor{black}{\textbf{4.3}} & \textbf{4.2} \\
    \bottomrule
    \end{tabular}
    }
    \label{rsp:dr_weather}
\end{minipage}
\hfill
\begin{minipage}{0.48\linewidth}
    \centering
    \caption{Zero-shot Non-Lambertian generalization. All models are trained on SceneFlow and evaluated using officially released weights.}
    \vspace{-0.2cm}
    \setlength{\tabcolsep}{1mm}{
    \begin{tabular}{c|c|ccccc}
    \toprule
    \multirow{2}{*}{Model} & Venue & $>$2px & $>$4px & $>$6px & $>$8px & Avg. (px) \\
    & & (\%) & (\%) & (\%) & (\%) & \\
    \midrule
    RAFTStereo~\cite{Lipson2021RAFTStereoMR} & 3DV'21 & 17.8 & 13.1 & 10.8 & 9.24 & 3.60 \\
    DLNR~\cite{zhao2023high} & CVPR'23 & 18.6 & 14.6 & 12.6 & 11.2 & 3.97 \\
    IGEV-Stereo~\cite{xu2023iterative} & CVPR'23 & 16.9 & 13.2 & 11.4 & 10.2 & 3.94 \\
    Selective-IGEV~\cite{wang2024selective} & CVPR'24 & 18.5 & 14.2 & 12.1 & 10.8 & 4.38 \\
    NRMF~\cite{guan2024neural} & CVPR'24 & 27.1 & 19.1 & 15.4 & 13.2 & 5.00 \\
    Former-CFNet$^\ddag$~\cite{zhanglearning2024} & ECCV'24 & 12.0 & 7.3 & 5.1 & 4.4 & 3.1 \\
    DEFOMStereo$^\ddag$~\cite{jiang2025defom} & CVPR'25 & \textcolor{black}{\textbf{11.6}} & \textcolor{black}{\textbf{7.1}} & \textcolor{black}{\textbf{4.7}} & \textcolor{black}{\textbf{4.1}} & \textcolor{black}{\textbf{2.9}} \\
    \midrule
    \textbf{SMFormer} & - & \textbf{11.4} & \textbf{7.0} & \textbf{4.5} & \textbf{4.0} & \textbf{2.8} \\
    \bottomrule
    \end{tabular}
    }
    \label{rsp:tab_booster}
\end{minipage}
\vspace{-0.2cm}
\end{table*}

\noindent\textbf{Middlebury~\cite{middlebury2014} \& ETH3D~\cite{Schps2017AMS}} 
The Middlebury dataset~\cite{middlebury2014} comprises 15 indoor training and 15 testing stereo pairs evaluated at half-resolution, using randomly cropped $320 \times 832$ patches with a maximum disparity of 320, while ETH3D~\cite{Schps2017AMS} contains 27 grayscale training and 20 testing pairs processed with $320 \times 640$ patches with a maximum disparity of 64.


\noindent\textbf{Booster~\cite{ramirez2022open}} comprises 228 training samples and 191 samples for online testing, spanning 64 distinct scenes with dense ground-truth disparity data. The dataset primarily features challenging non-Lambertian surfaces. In our experiments, we utilize the quarter-resolution version.
Since we use only the balanced training subset and thus only compare the ``balanced'' results on Booster benchmark~\cite{ramirez2022open}.
Specifically, We randomly cropped patches of size $320 \times 832$ for training.
The maximum disparity range is set to 192.

\begin{table*}[ht]
\begin{minipage}{\textwidth}
    \begin{minipage}[t]{0.50\textwidth}
    \caption{Quantitative evaluation on ETH3D and Middlebury \\benchmarks. ``H" denotes the half-resolution.}
    \vspace{-0.2cm}
    \centering
    \scriptsize
    \setlength{\tabcolsep}{1.5mm}{
    \begin{tabular}{l|c|cc|cc}
    \toprule
    \multirow{2}{*}{Model} &\multirow{2}{*}{\textbf{Supervised.}} &\multicolumn{2}{c|}{\textbf{ETH3D}} &\multicolumn{2}{c}{\textbf{Middle}} \\
    & & \textbf{Bad\ 1.0} & \textbf{AvgErr} & \textbf{Bad\ 2.0} & \textbf{AvgErr} \\
    \midrule
    ACVNet (H)~\cite{xu2022attention} & \Checkmark & \textbf{2.58} & \textbf{0.23} & 13.6 & 8.24 \\
    UPFNet (H)~\cite{chen2023unambiguous} &\Checkmark  & 3.82 & 0.25 & \textbf{10.3} & \textbf{1.90} \\
    AANet++ (H)~\cite{xu2020aanet} & \Checkmark  & 5.01 & 0.31 & 15.4 & 6.37 \\
    \midrule
    \textbf{SMFormer} (H)& \XSolidBrush  & \textbf{\textcolor{black}{3.54}} & \textbf{\textcolor{black}{0.25}} & \textbf{\textcolor{black}{12.8}} & \textbf{\textcolor{black}{4.37}} \\   
    \bottomrule
    \end{tabular}}
    \vspace{-0.3cm}
    \label{sec4:table_middle}
     \end{minipage}
    \begin{minipage}[t]{0.45\textwidth}
    \centering
    \scriptsize
    \caption{Benchmark results of the Booster dataset.  
    Res. denotes the image resolution. Note that we report ``All" results.}
    \vspace{-0.2cm}
    \setlength{\tabcolsep}{1.mm}{
    \begin{tabular}{l|c|c|cccc}
    \toprule
    Method & \textbf{Res.} & \textbf{Supervised.} & \textbf{Bad 2} & \textbf{Bad 4} & \textbf{Bad 8} & \textbf{MAE} \\
    \midrule
    RAFT-Stereo~\cite{Lipson2021RAFTStereoMR} & Quarter &  \Checkmark & 14.46 & 9.47 & 5.76 & 1.87\\	  
      RAFT-Stereo (32 iters)~\cite{Lipson2021RAFTStereoMR} & Quarter &  \Checkmark & \textbf{\textcolor{black}{10.73}} & \textbf{\textcolor{black}{6.79}} & \textbf{\textcolor{black}{3.87}} & \textbf{\textcolor{black}{1.29}}\\					
      CFNet~\cite{shen2021cfnet} & Quarter &  \Checkmark   & 29.65 & 19.94  & 12.88  & 4.90 \\
    CREStereo~\cite{li2022practical} & Quarter & \Checkmark  & \textbf{9.00} & \textbf{5.30} & \textbf{2.82} & \textbf{1.25} \\
      \midrule
      \textbf{SMFormer (Ours)} & Quarter &\XSolidBrush & 14.88  & {9.38} & {5.38} & {2.19} \\
      \bottomrule
    \end{tabular}}
    \label{sec4:booster}
    \vspace{-0.3cm}
\end{minipage}
\end{minipage}
\end{table*}

\begin{figure*}[t]
    \includegraphics[width=1.0\linewidth]{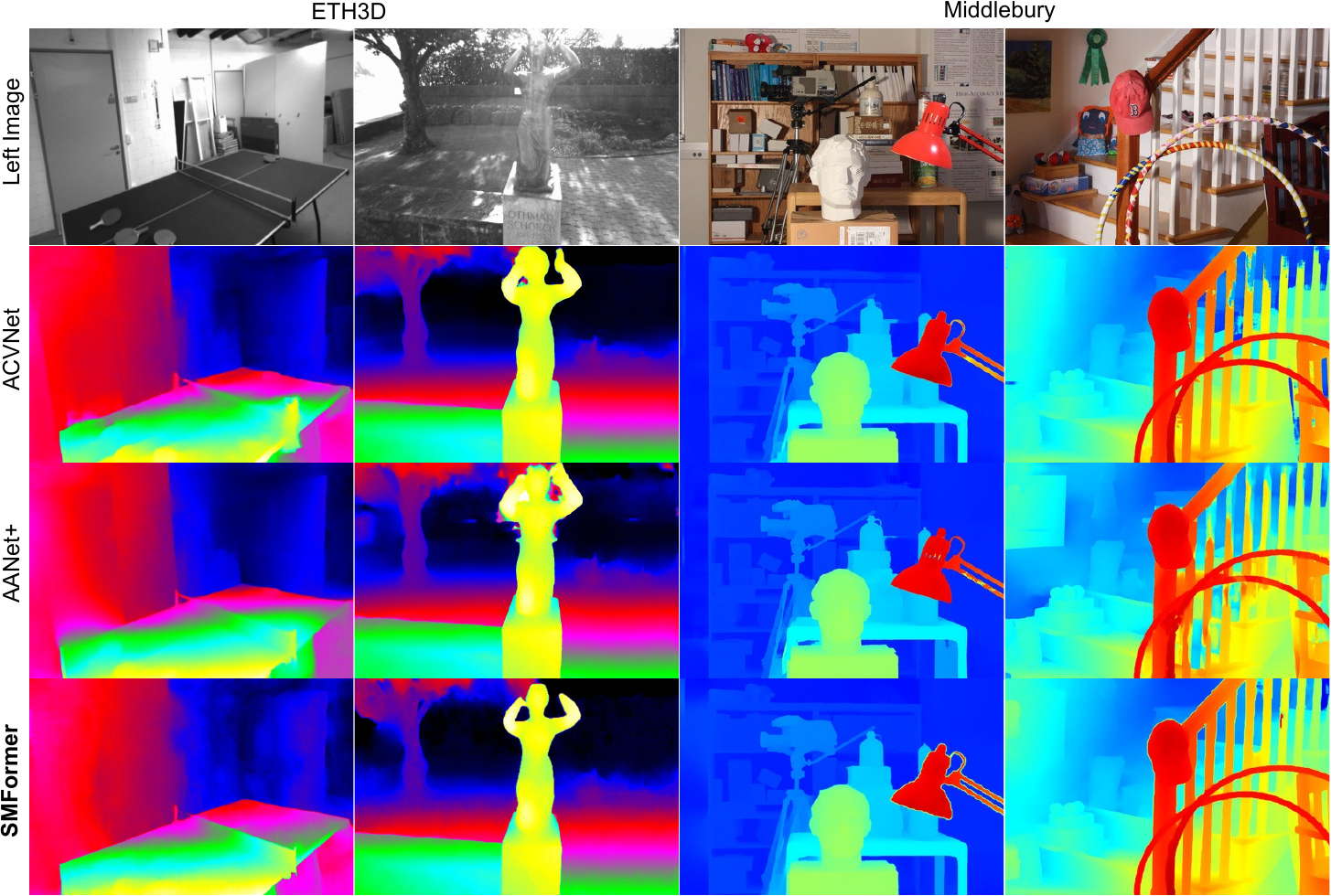}
    \vspace{-0.6cm}
    \caption{ Visualization results achieved by our method and other supervised methods. Disparity maps are generated from Middlebury \& ETH3D benchmarks.}
    \label{sec4:vis_comparison}
    \vspace{-0.3cm}
\end{figure*}

\noindent\textbf{Evaluation Metric.}
For evaluation metrics, end-point error (EPE) and t-pixel error rate ( Bad t) are adopted.
In addition, the percentage of stereo disparity outliers, defined as disparities larger than 3 pixels or more than 5\% of the ground truth disparities (referred to as D1), is utilized as a key metric. We also measure the average pixel error rate across different regions, including background (bg), foreground (fg), non-occluded regions (Noc), and all regions (All), to comprehensively evaluate the performance of different methods.

\subsubsection{Visual Foundation Models}
We use five distinct VFMs, each trained using varying strategies and datasets. 
The VFMs include DUST3R~\cite{dust3r_cvpr24}, the first 3D reconstruction pipeline, DINOV2~\cite{oquab2024dinov2} based on self-supervised pretraining with a curated dataset, EVA02~\cite{fang2023eva}, which integrates CLIP~\cite{clip2021} with masked image modeling, SAM~\cite{kirillov2023segment}, which capitalizes on a large-scale segmentation dataset, and DAMV2~\cite{yang2024depth_v2}, which collects and annotates large-scale unlabeled video data.

\begin{figure*}[t]
    \includegraphics[width=1.0\linewidth]{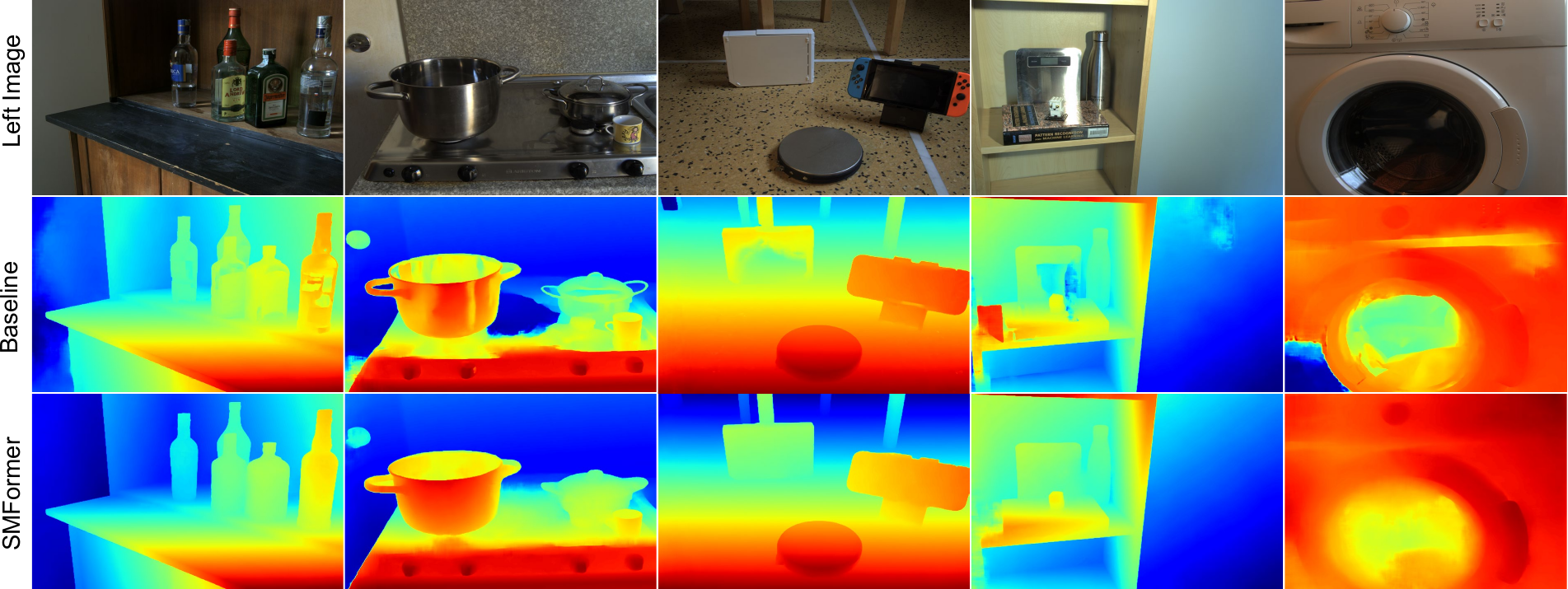}
    \vspace{-0.6cm}
    \caption{Visualization results achieved by our method and the baseline method. Note that the baseline method uses the CFNet backbone and employs the vanilla photometric loss and disparity smooth loss as supervisory signals.}
    \label{sec4:booster_comparison}
    \vspace{-0.4cm}
\end{figure*}

\begin{table}[t]
\centering
\scriptsize
\caption{Ablation study on different VFMs on zero-shot generalization. Note that all model variants are only trained on the synthetic SceneFlow dataset and tested on four real datasets. $\ast$ indicates that the officially provided weights are used for evaluation. And VFMs include the proposed MLA module.}
\vspace{-0.2cm}
\setlength{\tabcolsep}{.2mm}{
    \begin{tabular}{l|c|cc|cc|cc|cc}
    \toprule
    \multirow{2}{*}[-1pt]{VFMs} & Model & \multicolumn{2}{c|}{\textbf{KIT 2012 (All)}} & \multicolumn{2}{c|}{\textbf{KIT 2015 (All)}} & \multicolumn{2}{c|}{\textbf{MID (Noc)}} & \multicolumn{2}{c}{\textbf{ETH3D (Noc)}}\\
     & Capacity &  \textbf{EPE} &  \textbf{Bad 3.0}  & \textbf{EPE} &  \textbf{Bad 3.0 } & \textbf{EPE } & \textbf{Bad 2.0} & \textbf{EPE } & \textbf{Bad 1.0 }  \\
    \midrule
    \multirow{1}{*}[-1pt]{Baseline${^\ast}$~\cite{shen2021cfnet}} & \XSolidBrush & 1.10 & 4.73 & 1.46 & 5.78 & 3.21 & 15.4 & 0.56 & 5.80  \\
    \midrule
    DUST3R~\cite{dust3r_cvpr24} & Large & 0.92 & 4.67  & 1.01 & 5.06 & 1.37 & 9.45 & 0.37 & 4.67 \\
    \midrule
     \multirow{2}{*}{DINOV2~\cite{oquab2024dinov2}} & Base & 0.93 & 4.56 & 1.02 & 5.14 & 1.41 & 9.79 & 0.37 & 3.93 \\
      & Large & 0.87 & 4.35 & $\textbf{0.96}$ & 4.92 & 1.29 & 8.61 & 0.34 & 3.49 \\
    \midrule
    \multirow{2}{*}{EVA2~\cite{fang2023eva}} & Base & 0.94 & 4.46 & 1.04 & 5.10 & 1.55 & 10.4 & 0.36 & 3.71 \\
    & Large & 0.88 & 4.27 & 1.01 & 4.98 & 1.40 & 9.83 & $\textbf{\textcolor{black}{0.33}}$ & {3.36} \\
    \midrule
    \multirow{2}{*}{DAMV2~\cite{depthanything}} & Base & 0.89 & 4.37 & 1.05 & 5.01 & 1.31 & 8.72 & 0.34 & 3.30  \\
      & Large & $\textbf{\textcolor{black}{0.83}}$ & $\textbf{\textcolor{black}{4.20}}$ & $$\textbf{\textcolor{black}{0.97}}$$ & $\textbf{\textcolor{black}{4.73}}$ & $\textbf{\textcolor{black}{1.22}}$ & $\textbf{\textcolor{black}{8.34}}$ & $\textbf{0.31}$ & \textcolor{black}{\textbf{2.97}} \\
      \midrule
    \multirow{2}{*}{SAM~\cite{kirillov2023segment}} & Base & 0.88 & 4.25 & 1.03 & 4.98 & 1.26 & 8.49 & 0.33 & 3.24 \\
      & Large & \textbf{0.82} & \textbf{4.12} & {0.98} & \textbf{4.71} & \textbf{1.20} & \textbf{8.13} & \textbf{0.31} & \textbf{2.89}  \\
    \bottomrule
    \end{tabular}}
\label{sec4:vfm}
\vspace{-0.4cm}
\end{table}

\subsubsection{Training Details}
SAM~\cite{kirillov2023segment} (ViT-Large) and CFNet~\cite{shen2021cfnet} are adopted as our backbone.
We first train our network on the synthetic SceneFlow dataset~\cite{dosovitskiy2015flownet} in a supervised manner and perform generalization evaluation on KITTI, Middlebury, and ETH3D training datasets to provide reasonable parameters for fine-tuning.
The initial learning rate is set to $1\times10^{-3}$ for 10 epochs and decreased to $1\times10^{-4}$ for another 5 epochs.
Then, we finetune our network on KITTI~\cite{geiger2012we,menze2015object}, Middlebury~\cite{middlebury2014},  ETH3D~\cite{Schps2017AMS}, and Booster~\cite{ramirez2022open} training datasets in a self-supervised learning manner to obtain the final model for submission.
The initial learning rate is set to $1\times10^{-4}$ for 40 epochs and decreased to $1\times10^{-5}$ for another 160 epochs. 
All models are optimized using the Adamw optimizer with $\beta_{1}$= 0.9 and $\beta_{2}$ = 0.999 and a batch size of 8. 
We use 8 NVIDIA 5000 Ada GPUs for all training experiments.



\subsection{Comparison to SOTA Methods}

\noindent \textbf{Cross-domain Generalization.}
We conduct experiments to demonstrate that our model, pre-trained on the synthetic SceneFlow dataset~\cite{dosovitskiy2015flownet}, achieves robust zero-shot generalization on real-world datasets. 
\textcolor{black}{As shown in Table~\ref{sec4:dg_comparison}, our SMFormer outperforms most domain generalization methods without relying on any tailored strategies. Additionally, as shown in Table~\ref{rsp:dr_weather} and Table~\ref{rsp:tab_booster}, SMFormer achieves strong performance under challenging real-world conditions, including the DrivingStereo Weather subset\cite{chen2015deepdriving} and the Booster dataset\cite{zamaramirez2022booster}. Notably, it outperforms recent open-source VFM-based methods such as Former-CFNet\cite{zhanglearning2024} and DEFOMStereo\cite{jiang2025defom}, highlighting its superior generalization capabilities.
These results confirm that SMFormer not only generalizes effectively across standard benchmarks (KITTI 2012~\cite{geiger2012we}, KITTI 2015~\cite{menze2015object}, Middlebury~\cite{middlebury2014}, and ETH3D~\cite{Schps2017AMS}) but also exhibits strong robustness in diverse challenging environments.}


\noindent\textbf{KITTI.}
To make a fair comparison, following the standard fine-tuning setup~\cite{wang2020parallax,li2018occlusion,xu2020aanet}, we use a mixture of KITTI 12 \& 15 (394 image pairs) for training. As shown in Table~\ref{sec4:table_kitti}, our method outperforms other up-to-date self-supervised methods by notable margins and even surpasses several supervised ones, except for ChiT~\cite{su2022chitransformer}. ChiT uses KITTI eigen splits (22600 image pairs, 57 times more than ours), resulting in comparable performance. Larger data capacity leads to better performance, as shwon in Table~\ref{sec4:capacity}.
Besides, the qualitative results verify the advantages of {SMFormer} shown in Fig.~\ref{sec4:vis_kit}. 


\noindent\textbf{Middlebury \& ETH3D.}
To our knowledge, prior self-supervised approaches~\cite{su2022chitransformer,wang2020parallax,zhang2019dispsegnet,Unos2019,liu2020flow2stereo} appear not to have published results on the Middlebury and ETH3D benchmarks. 
This may be due to the complexity of these benchmarks, containing many image pairs that violate the photometric consistency assumption.
We also validate that using the photometric consistency loss indeed results in poor performance on Middlebury and ETH3D (Table~\ref{sec4:losses}, 1st row).
Therefore, we compare only mainstream supervised methods, which handle challenging regions better due to ground truth. 
Surprisingly, SMFormer outperforms several supervised methods, including AANet~\cite{xu2020aanet} on ETH3D and Middlebury benchmarks, and surpasses ACVNet~\cite{xu2022attention} on Middlebury (Table~\ref{sec4:table_middle}). 
Fig.~\ref{sec4:vis_comparison} presents a qualitative comparison of the disparity maps estimated by our method and those produced by mainstream supervised methods.
Overall, this fully demonstrates SMFormer's superiority and ability to handle extremely challenging areas without any ground truth supervision. 

\begin{figure*}[t]    
    \includegraphics[width=1.0\linewidth]{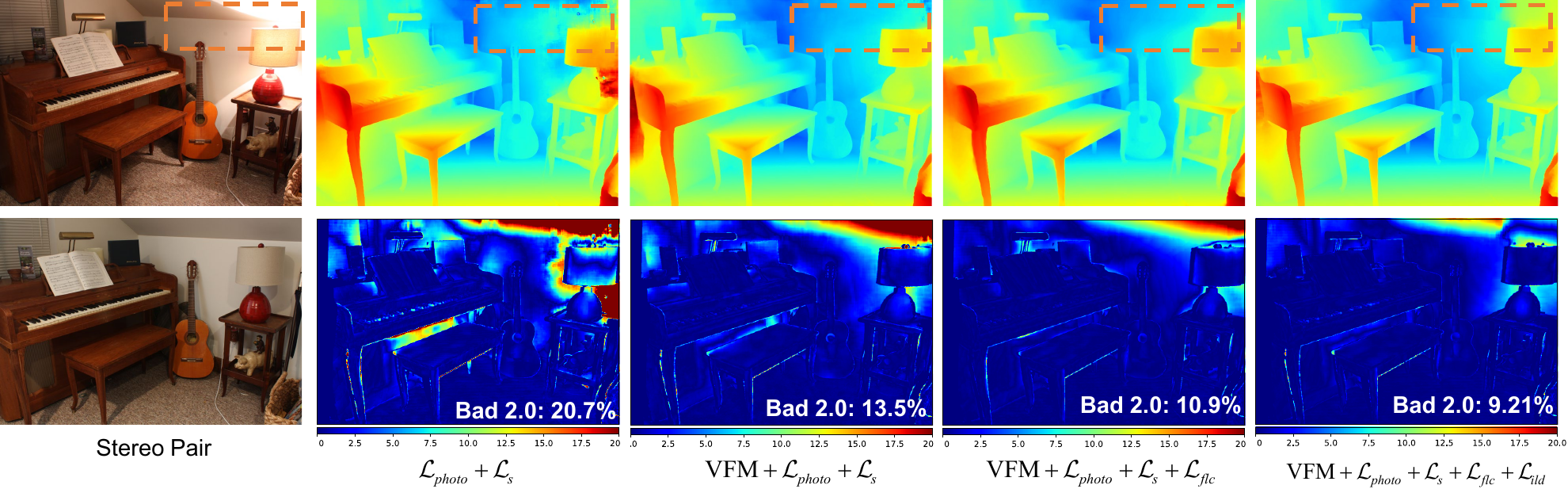}
    \vspace{-0.6cm}
    \caption{Visual comparison of SMFormer on Middlebury PianoL pair. Top row: the predicted disparity maps with different combinations of the proposed components. Bottom row: the error maps. 
     $\mathcal{L}_{photo}$ is the photometric consistency loss;  $\mathcal{L}_{s}$ is the disparity smooth loss. 
    $\mathcal{L}_{flc}$: The feature-level Stereo Contrastive Loss; $\mathcal{L}_{ild}$: The image-level Disparity Difference Loss.
    }
    \vspace{-0.3cm}
\label{sec4:middle_vis}
\end{figure*}

\begin{table*}[t]
\centering
\scriptsize
\caption{Ablation study on the proposed components. We evaluate the model on the KITTI 2012 \& 2015, Middlebury, and ETH3D training sets, using all values (All) rather than just the non-occluded regions (Noc). VFM denotes SAM~\cite{kirillov2023segment} (ViT-Large).
\textcolor{cyan}{Cyan} color in the table represents the results of the baseline, while \textcolor{pink}{pink} color represents the results of the final model.}
\vspace{-0.2cm}
\setlength{\tabcolsep}{2.5mm}{
    \begin{tabular}{c|c|cccc|cc|cc|cc|cc}
    \toprule
    \multirow{2}{*}[-1pt]{Model} & \multirow{2}{*}[-1pt]{\textbf{Feature Extractor}} & \multirow{2}{*}[-1pt]{$\mathcal{L}_{photo}$} & \multirow{2}{*}[-1pt]{$\mathcal{L}_{s}$} & \multirow{2}{*}[-1pt]{$\mathcal{L}_{flc}$} & \multirow{2}{*}[-1pt]{$\mathcal{L}_{ild}$} & \multicolumn{2}{c|}{\textbf{KIT 2012 (All)}} & \multicolumn{2}{c|}{\textbf{KIT 2015 (All)}} & \multicolumn{2}{c|}{\textbf{Middle (All)}} & \multicolumn{2}{c}{\textbf{ETH3D (All)}}\\
    & & & & & & \textbf{EPE} &  \textbf{D1}  & \textbf{EPE} &  \textbf{D1}  & \textbf{EPE} & \textbf{Bad 2.0}  & \textbf{EPE} & \textbf{Bad 1.0}   \\
    \midrule
    Baseline (CFNet) & FPN & \Checkmark & \Checkmark &  &  &  \cellcolor{cyan} 0.84 & \cellcolor{cyan} 4.10 & \cellcolor{cyan} 0.96  & \cellcolor{cyan} 4.27 & \cellcolor{cyan} 2.87 & \cellcolor{cyan} 13.8 & \cellcolor{cyan} 1.62 & \cellcolor{cyan} 10.2 \\
    \midrule
    \multirow{4}{*}{\textbf{SMFormer}} & VFM & \Checkmark & \Checkmark & & & 0.87 & 4.56 & 1.01  & 4.83 & 1.74 & 11.3 & 1.27 &  9.24 \\
    & FPN + VFM & \Checkmark & \Checkmark & & & 0.77 & 3.98 & 0.85 & 4.05 & 1.29 & 9.81 & 0.49 & 5.25\\
    & FPN + VFM & \Checkmark & \Checkmark & \Checkmark & & \underline{0.73} & \underline{3.59} & \underline{0.82} & \underline{3.67} &  \underline{1.10} & \underline{8.86} & \underline{0.36} & \underline{3.55}  \\
     & FPN + VFM & \Checkmark & \Checkmark & \Checkmark & \Checkmark & \cellcolor{pink} $\textbf{0.68}$ & \cellcolor{pink} $\textbf{3.01}$ & \cellcolor{pink} $\textbf{0.76}$ & \cellcolor{pink} $\textbf{3.10}$ & \cellcolor{pink} $\textbf{0.95}$ & \cellcolor{pink} $\textbf{8.14}$ & \cellcolor{pink} $\textbf{0.28}$ & \cellcolor{pink} $\textbf{2.93}$ \\
    \bottomrule
    \end{tabular}}
    \vspace{-0.3cm}
\label{sec4:losses}
\end{table*}

\begin{table}[t]
\centering
\scriptsize
\caption{\textcolor{black}{Ablation studies with different components on Feature extractor using the baseline losses ($\mathcal{L}_{photo}$ and $\mathcal{L}_{s}$). Following~\cite{zhanglearning2024}, we adopt a DPT decoder strategy to act as the adaptor.}}
\vspace{-0.2cm}
\setlength{\tabcolsep}{0.5mm}{
    \begin{tabular}{cc|cc|cc|cc|cc}
    \toprule
   \multicolumn{2}{c|}{\textbf{Feature Extractor}} & \multicolumn{2}{c|}{\textbf{KIT 2012}} & \multicolumn{2}{c|}{\textbf{KIT 2015}}  & \multicolumn{2}{c|}{\textbf{Middle}} & \multicolumn{2}{c}{\textbf{ETH3D}}\\
    Backbone & MLA & \textbf{EPE} &  \textbf{D1}  & \textbf{EPE} &  \textbf{D1}  & \textbf{EPE} & \textbf{Bad 2.0} & \textbf{EPE} & \textbf{Bad 1.0} \\
    \midrule
    SAM & \XSolidBrush & 0.87 & 4.56 & 1.01  & 4.83 & 1.74 & 11.3 & 1.27 &  9.24 \\
    SAM + Adapter~\cite{zhanglearning2024} & \XSolidBrush & 0.84 & 4.15 & 0.93 & 4.26 & 1.49 & 10.4 & 0.62 & 6.86 \\
    SAM + FPN & \XSolidBrush & \underline{0.81} & \underline{4.04} & \underline{0.92} & \underline{4.13} & \underline{1.37} & \underline{10.2} & \underline{0.55} & \underline{5.79} \\
    SAM + FPN & \Checkmark & \textbf{0.77} & \textbf{3.98} & \textbf{0.85} & \textbf{4.05} & \textbf{1.29} & \textbf{9.81} & \textbf{0.49} & \textbf{5.25} \\
    \bottomrule
    \end{tabular}}
    \vspace{-0.3cm}
\label{sec4:msa}
\end{table}

\noindent\textbf{Booster.}
The Booster benchmark features highly reflective and textureless surfaces that violate photometric consistency, posing significant challenges for self-supervised stereo matching methods in these regions. Since prior self-supervised approaches lack submissions on Booster, we benchmark against recent supervised methods (e.g., RAFT-Stereo~\cite{Lipson2021RAFTStereoMR}, CFNet~\cite{shen2021cfnet}), which leverage ground truth data to better address such challenges. As shown in Table~\ref{sec4:booster}, our method surpasses CFNet significantly even without the supervision of ground truth. Qualitative results in Fig.~\ref{sec4:booster_comparison} further demonstrate our approach’s robustness in reflective and textureless areas, underscoring its effectiveness.



\subsection{Ablation Study}
\begin{figure}[t]
    \scriptsize
    \includegraphics[width=1.0\linewidth]{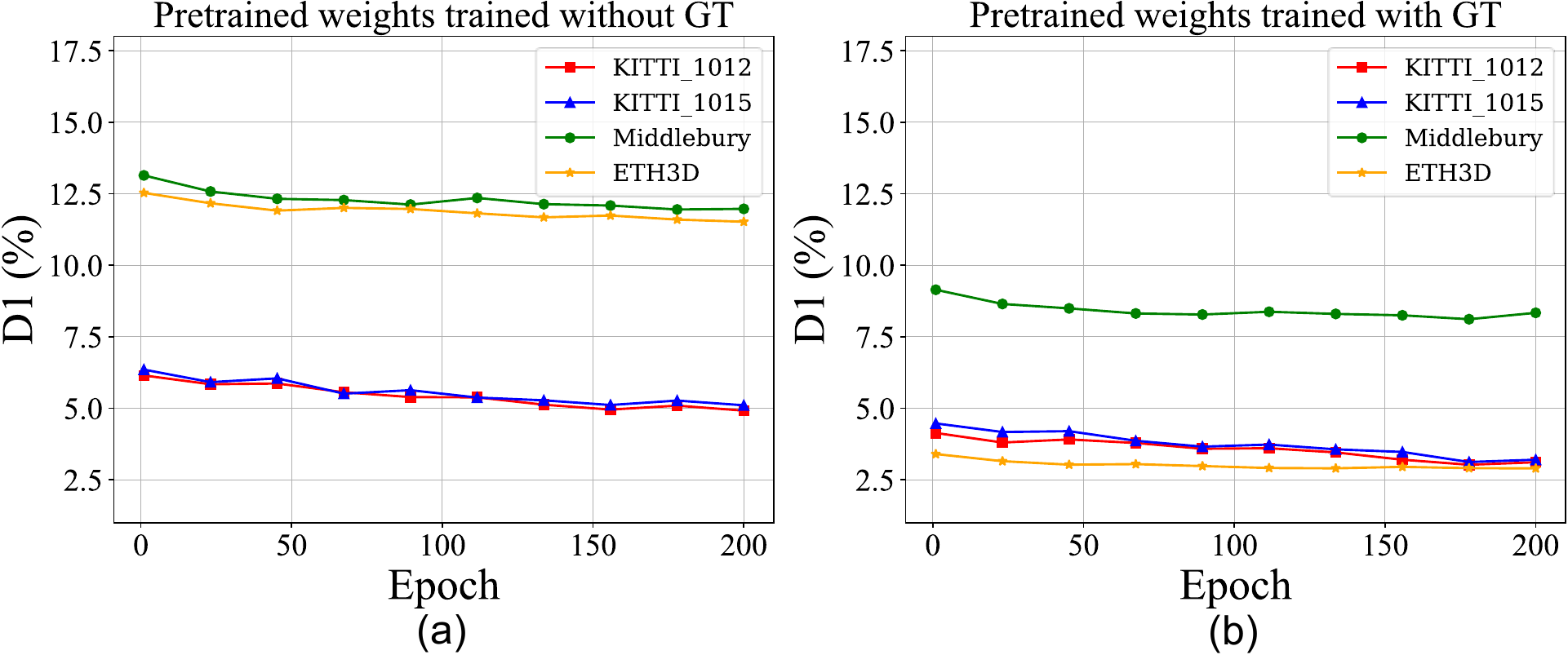}
    \vspace{-0.5cm}
    \caption{Comparison with different pre-trained weights on KITTI 2012, KITTI 2015, Middlebury, and ETH3D datasets. D1 denotes KITTI (Bad 3.0), Middlebury (Bad 2.0), and ETH3D (Bad 1.0) metrics.}
    \label{sec4:d1_curve}
    \vspace{-0.4cm}
\end{figure}

\textcolor{black}{Across Tables~\ref{sec4:losses}, \ref{sec4:msa}, and \ref{sec4:da}, we verify that our design is effective for self-supervised stereo learning, especially in ill-posed regions where vanilla photometric-consistency supervision is unreliable. The main gain comes from integrating VFM representations with conventional pyramid features (FPN+VFM), which improves robustness under challenging appearances and domain shifts. On top of this hybrid representation, the proposed loss terms provide complementary benefits by strengthening cross-view consistency and enhancing robustness in ill-posed regions, thereby stabilizing self-supervision under challenging conditions (e.g., illumination changes, reflections, and texture-less areas). We further analyze practical factors including pre-trained initialization (Fig.~\ref{sec4:d1_curve}),   VFM choices (Table~\ref{sec4:vfms}), hyperparameter sensitivity (Table~\ref{sec4:occ} and \ref{sec4:hyper}), training data scale (Table~\ref{sec4:capacity}), and training cost (Table~\ref{supp:cost}). Overall, these studies demonstrate that the full model consistently outperforms the vanilla photometric-consistency baseline. Below we present detailed results.}

\noindent\textbf{Different VFMs.}
From Table~\ref{sec4:vfm}, we observe two key points:
1) VFMs consistently enhance baseline generalization performance. For instance, compared to the robust baseline CFNet~\cite{shen2021cfnet}, EPE of {SMFormer} (with SAM) is reduced by 25.5\%, 32.8\%, 63.2\%, and 46.5\%, respectively.
2) Larger model capacity enhances zero-shot performance due to stronger feature representation and robust priors.
Therefore,
we adopt SAM~\cite{kirillov2023segment} (ViT-Large) as the feature extractor backbone.

\begin{table}[t]
\centering
\scriptsize
\caption{Ablation study on different data augmentation strategies. ``Vanilla'' and ``Intermediate'' use the augmented pair as input in a single standard branch and the baseline losses ($\mathcal{L}_{photo}$ and $\mathcal{L}_{s}$), where $\mathcal{L}_{photo}$ uses augmented (Vanilla) or clean (Intermediate) image pair. In contrast, our method adopts an additional data augmentation branch and the proposed losses ($\mathcal{L}_{flc}$ and $\mathcal{L}_{ild}$).
}
\vspace{-0.2cm}
\setlength{\tabcolsep}{1mm}{
    \begin{tabular}{l|c|cc|cc|cc|cc}
    \toprule
    \multirow{2}{*}[-1pt]{ID} & Data &  \multicolumn{2}{c|}{\textbf{KIT 2012}} & \multicolumn{2}{c|}{\textbf{KIT 2015}}  & \multicolumn{2}{c|}{\textbf{Middle}} & \multicolumn{2}{c}{\textbf{ETH3D}}\\
    & Augmentation & \textbf{EPE} &  \textbf{D1}  & \textbf{EPE} &  \textbf{D1}  & \textbf{EPE} & \textbf{Bad 2.0} & \textbf{EPE} & \textbf{Bad 1.0} \\
    \midrule
    1 & \XSolidBrush & 0.77 & 3.98 & 0.85 & 4.05 & 1.29 & 9.81 & 0.49 & 5.25 \\
    2 & Vanilla & 0.81 & 4.23 & 0.90 & 4.36 & 1.96 & 13.1 & 0.81 & 9.65 \\
    3 & Intermediate & \underline{0.74} & \underline{3.75} & \underline{0.83} & \underline{3.87} & \underline{1.11} & \underline{9.10} & \underline{0.41} & \underline{4.58} \\
    4 & \textbf{Ours} & \textbf{0.68} & \textbf{3.01} & \textbf{0.76} & \textbf{3.10} & \textbf{0.95} & \textbf{8.14} & \textbf{0.28} & \textbf{2.93} \\
    \bottomrule
    \end{tabular}}
\label{sec4:da}
\vspace{-0.2cm}
\end{table}

\noindent\textbf{The Proposed Components.}
\textcolor{black}{Table~\ref{sec4:losses} provides a stepwise ablation of the feature extractor and the proposed self-supervised objectives. Overall, the largest and most consistent gain comes from combining FPN with VFM features (FPN+VFM), especially on challenging benchmarks such as Middlebury and ETH3D, indicating improved robustness in reflective and texture-less regions. On top of this hybrid representation, $L_{\text{flc}}$ and $L_{\text{ild}}$ provide complementary gains: $L_{\text{flc}}$ strengthens feature-level stereo consistency, while $L_{\text{ild}}$ promotes context-aware robustness against common disturbances, thereby stabilizing self-supervised supervision. We therefore include all components in the final model.
Fig.~\ref{sec4:middle_vis} further corroborates these findings qualitatively: on the PianoL example with strong illumination changes and large reflective/texture-less areas, training with only the vanilla losses ($\mathcal{L}_{photo}$ and $\mathcal{L}_{s}$) produces blurrier disparity maps, whereas adding the proposed components leads to sharper and more coherent predictions. Overall, both quantitative and qualitative results validate the effectiveness of our design.}

\noindent\textbf{\textcolor{black}{Analysis of MLA.}}
\textcolor{black}{From Table~\ref{sec4:msa}, our model benefits from the learnable FPN network and the MLA mechanism. Additionally, combining the FPN network with VFM is more effective than using the Adapter~\cite{zhanglearning2024} to retrieve target domain information, leading to better performance. Overall, FPN extracts rich target domain information, while VFM with MLA mechanism captures global context across views and layers, enhancing robust feature representation for stereo matching.}

\noindent\textbf{Data Augmentation Strategies.}
\textcolor{black}{Table~\ref{sec4:da} compares different augmentation schemes for {SMFormer}. 
Applying ``vanilla'' augmentation to the stereo pair (ID=2) degrades performance since it violates the photometric-consistency assumption of $\mathcal{L}_{photo}$. 
The ``intermediate'' scheme (ID=3) brings the major improvement by restoring valid photometric supervision, while our final strategy (ID=4) provides further gains by improving robustness to appearance disturbances under reliable self-supervised constraints. 
Overall, preserving photometrically consistent supervision is the primary contributor, and our augmentation offers complementary robustness benefits.}


\begin{table}[t]
\centering
\scriptsize
\caption{\textcolor{black}{Performance comparison with the proposed SMFormer across multiple VFMs as backbones. We incorporate the ViT-Large architectures of VFMs into our framework while keeping the other proposed components and training strategy unchanged.}}
\setlength{\tabcolsep}{1.5mm}{
    \begin{tabular}{l|cc|cc|cc|cc}
    \toprule
    \multirow{2}{*}[-1pt]{VFMs} & \multicolumn{2}{c|}{\textbf{KIT 2012}} & \multicolumn{2}{c|}{\textbf{KIT 2015}}  & \multicolumn{2}{c|}{\textbf{Middle}} & \multicolumn{2}{c}{\textbf{ETH3D}}\\
    & \textbf{EPE} &  \textbf{D1}  & \textbf{EPE} &  \textbf{D1}  & \textbf{EPE} & \textbf{Bad 2.0} & \textbf{EPE} & \textbf{Bad 1.0} \\
    \midrule
    DINOV2~\cite{oquab2024dinov2} & 0.72 & 3.27 & 0.79 & 3.39 & 1.27 & 8.96 & 0.40 & 4.29 \\
    EVA2~\cite{fang2023eva} & 0.71 & 3.30 & 0.81 & 3.46 & 1.30 & 9.21 & 0.35 & 4.10 \\
    DAM~\cite{depthanything}& \textbf{0.66} & \textbf{3.02} & 0.78 & {3.18} & {1.01} & {8.35} & {0.29} & {3.24} \\ 
    SAM~\cite{kirillov2023segment}  & {0.68} & {3.01} & \textbf{0.76} &  \textbf{3.10} & \textbf{0.95} & \textbf{8.14} & \textbf{0.28} & \textbf{2.93} \\
    \bottomrule
    \end{tabular}}
\label{sec4:vfms}
\vspace{-0.2cm}
\end{table}
\noindent\textbf{The Impact of Pre-trained Weights.}
Fine-tuning pre-trained models to the target domain has been ubiquitously proven effective in stereo matching.
Previous self-supervised stereo methods~\cite{wang2020parallax, li2018occlusion} are usually pre-trained on SceneFlow in a self-supervised manner rather than a supervised manner to provide reasonable weights.
We posit that supervised pre-training on synthetic data can enhance robustness by leveraging GT-derived strong supervision signals, which improve occlusion and reflective region handling. To validate this, we train SMFormer on SceneFlow in both self-supervised (Fig.~\ref{sec4:d1_curve} (a)) and supervised (Fig.~\ref{sec4:d1_curve} (b)) configurations, then fine-tune on real-world datasets in a self-supervised manner. Empirical results show the supervised pre-trained model achieves significantly superior performance, underscoring the value of GT-guided synthetic training for real-world generalization.


\noindent\textbf{\textcolor{black}{Study on Various VFMs.}}
\textcolor{black}{Our SMFormer exhibits compatibility with a range of vision foundation models, as shown in Table~\ref{sec4:vfms}. These model variants achieve state-of-the-art performance on these four distinct real-world datasets.
Notably, these results demonstrate the compatibility of SMFormer with different ViT pretraining backbones.
Since SAM~\cite{kirillov2023segment} achieves the best performance among the evaluated VFM variants, we adopt it as the final choice in our framework.}

\begin{table}[t]
\centering
\scriptsize
\caption{Ablation study of the occlusion rate $\alpha$ for the pixel level contrastive sample on the KITTI 2015 and Middlebury training sets.}
\vspace{-0.2cm}
    \setlength{\tabcolsep}{3mm}{
    \begin{tabular}{l|cc|cc}
    \toprule
    \multirow{2}{*}[-1pt]{\textbf{Occlusion Rate}} & \multicolumn{2}{c|}{\textbf{KIT 2015}} & \multicolumn{2}{c}{\textbf{Middle}} \\
    & \textbf{EPE} & \textbf{D1} & \textbf{EPE} & \textbf{Bad 2.0 }\\
    \midrule
     $\alpha$ = 0 &  0.84 & 3.86 & 1.19 & 9.45 \\
     $\alpha$ = 0.05  & {0.78} & {3.20} & {1.05} & {8.70} \\
    $\alpha$ = 0.15 & \textbf{0.76} & \textbf{3.10} & \textbf{0.95} & \textbf{8.14} \\
    $\alpha$ = 0.25 & {0.82} & {3.35} & {1.02} & {8.47} \\
    \bottomrule
    \end{tabular}}
\label{sec4:occ}
\vspace{-0.4cm}
\end{table}

\noindent\textbf{Occlusion Rate.}
As shown in Table~\ref{sec4:occ}, we study the pixel-level occlusion rate $\alpha$ used to construct contrastive samples. Overall, an appropriate $\alpha$ noticeably improves performance, indicating that moderate masking enhances accuracy and robustness. However, overly large masking ratios degrade performance for two reasons: (i) dense matching becomes excessively difficult, introducing noisy gradients that hinder disparity learning; and (ii) excessive masking reduces the effective matching area for subsequent modules (e.g., feature matching). Consequently, performance improves as $\alpha$ increases to a moderate level (around 15\%), but then drops for larger ratios. This trend is consistent with masked stereo matching~\cite{rao2023masked}, where higher masking ratios make reconstruction harder and shrink the valid matching region.

\begin{table}[t]
\centering
\scriptsize
\caption{\textcolor{black}{Ablation study of hyperparameters temperature $\tau$ and the threshold value $\tau_{warp}$ on the KITTI 2015 training sets.}}
\vspace{-0.2cm}
    \setlength{\tabcolsep}{3mm}{
    \begin{tabular}{l|cc|l|cc}
    \toprule
    \multirow{2}{*}[-1pt]{\textbf{Temperature}} & \multicolumn{2}{c|}{\textbf{KIT 2015}} & \multirow{2}{*}[-1pt]{\textbf{Threshold}}  & \multicolumn{2}{c}{\textbf{KIT 2015}} \\
    & \textbf{EPE} & \textbf{D1} & & \textbf{EPE} & \textbf{D1}\\
    \midrule
     $\tau $ = 1 (Uniform) &  0.80 & 3.46  & $\tau_{warp}$ = 1 & 0.79 & 3.35 \\
     $\tau$ = 0.1  & {0.77} & {3.18}  & $\tau_{warp}$ = 2 & {0.76} & {3.21} \\
    $\tau$ = 0.07 & \textbf{0.76} & \textbf{3.10} & $\tau_{warp}$ = 3 & \textbf{0.76} & \textbf{3.10} \\
    $\tau$ = 0.01 & {0.79} & {3.37} & $\tau_{warp}$ = 5 & {0.81} & {3.51} \\
    \bottomrule
    \end{tabular}}
\label{sec4:hyper}
\vspace{-0.3cm}
\end{table}


\begin{table}[t]
    \centering
    \scriptsize
    \caption{Performance Gains from Data Capacity. KIT, VKITTI2, and KITTI RAW denote the mixed KITTI 2012\& KITTI 2015, Virtual KITTI 2, and KITTI RAW datasets, respectively.}
    \vspace{-0.2cm}
    \setlength{\tabcolsep}{2mm}{
    \begin{tabular}{c|cc|cc}
    \toprule
     \multirow{2}{*}[-1pt]{\textbf{Training Dataset}} &  \multicolumn{2}{c|}{\textbf{KIT 2012}} & \multicolumn{2}{c}{\textbf{KIT 2015}} \\
    & \textbf{EPE} & \textbf{D1} & \textbf{EPE} & \textbf{D1} \\ 
    \midrule
    KIT & 0.68 & 3.01 & 0.76 & 3.10\\
    KIT + VKITTI2 & 0.61 & 2.68 & 0.70 & 2.79 \\
    KIT + VKITTI2 + KITTI RAW & \textbf{0.57}& \textbf{2.55} & \textbf{0.66} & \textbf{2.62} \\
    \bottomrule
    \end{tabular}
    \label{sec4:capacity}
    \vspace{-0.3cm}}
\end{table}

\begin{table}[t]
    \centering
    \scriptsize
    \caption{\textcolor{black}{The quantitative training time comparisons. ``Full" denotes full-finetuning. `S' denotes the single-branch baseline using vanilla photometric consistency loss, while `D' denotes the dual-branch setting used in our training strategy.}}
    \vspace{-0.2cm}
    \setlength{\tabcolsep}{2mm}{
    \begin{tabular}{c|c|c|c|c|c}
    \toprule
     \multirow{1}{*}[-1pt]{\textbf{VFM}}  & \multirow{1}{*}[-1pt]{\textbf{VFM}} & \multirow{2}{*}[-1pt]{\textbf{Method}}  &  \multicolumn{1}{c|}{\textbf{Training}} & \multicolumn{1}{c|}{\textbf{Number of}} &  \multicolumn{1}{c}{\textbf{Training}} \\
     \textbf{Model} & \textbf{Size} & & \textbf{Paradigm} & \textbf{Epochs} & \textbf{Time} \\
     \midrule
    \multirow{3}{*}[-1pt]{\textbf{SAM}} & ViT-Large & {Baseline} & S & 200 & 6.0 h \\
    & ViT-Large & Full & {D} & 200 & 16.9 h \\
    &  ViT-Large & \textbf{Ours} & \textbf{D} & 200 & 8.6 h \\
    \bottomrule
    \end{tabular}}
    \label{supp:cost}
    \vspace{-0.3cm}
\end{table}

\noindent\textcolor{black}{\textbf{Hyperparameter Analysis.}
Table~\ref{sec4:hyper} investigates the effects of the contrastive temperature $\tau$ and the warping threshold $\tau_{\text{warp}}$ on KITTI~2015. Overall, we observe that {moderate} hyperparameter choices consistently yield the best performance (EPE $=0.76$, D1 $=3.10$): a mid-range $\tau$ strikes a favorable balance between highlighting informative hard negatives and preserving stable optimization, whereas overly large or overly small values lead to suboptimal training. Likewise, an intermediate $\tau_{\text{warp}}$ provides the best trade-off between filtering out unreliable warps and retaining sufficient supervision, while thresholds that are too strict or too permissive degrade accuracy. We therefore use these moderate default settings for $\tau$ and $\tau_{\text{warp}}$ in all subsequent experiments.
}

\noindent\textbf{More Data Capacity.}
We investigate how training data volume affects SMFormer by incrementally adding datasets: the KITTI 12 \& 15 mixed set~\cite{geiger2012we,menze2015object}, KITTI RAW (Eigen Splits, 22600 pairs), and the synthetic Virtual KITTI 2 (20 $k$ samples)~\cite{cabon2020virtual}. As shown in Table~\ref{sec4:capacity}, performance on KITTI 2012 \& 2015 consistently improves with larger training data. To align with standard practice in stereo matching (where most methods use only mixed KITTI datasets for fine-tuning), we adopt identical settings for fair benchmarking.


\noindent\textbf{Training Costs.}
\textcolor{black}{We systematically benchmark the training cost to provide a comprehensive evaluation of the practical applications. As summarized in Table~\ref{supp:cost}, when trained on 320 $\times$ 832-resolution KITTI stereo pairs using a single RTX 5000 Ada GPU with a batch size of 1, the \textbf{S}ingle-branch baseline requires 6.0 hours to complete 200 epochs. Under the same configuration, the proposed Dual-branch framework completes training in just 8.6 hours, introducing only a modest $\sim$1.4$\times$ overhead despite its dual-branch design. This efficiency stems from our PyTorch-based implementation, where operations are scheduled sequentially, but the two branches serve as independent feature extraction paths. On modern GPUs, these computations are executed in parallel at the kernel level, preventing the overhead from scaling linearly with the number of branches. In contrast, full fine-tuning requires 16.9 hours of training, underscoring the favorable trade-off our framework achieves between computational efficiency and performance.}

\section{limitation}
\textcolor{black}{Our method leverages strong priors from large pre-trained vision foundation models, which may add computational overhead in practical deployments; lightweight backbones or distillation could help meet stricter real-time budgets. In addition, compared to standard self-supervised stereo, our approach introduces additional training components and multi-stage optimization, so careful hyperparameter tuning and staged schedules are often beneficial for ensuring consistently strong performance across datasets and domains.}

\section{Conclusion}
In this paper, we introduce SMFormer, a novel self-supervised stereo matching framework that bridges the performance gap caused by the coarse hypothesis of photometric consistency in previous self-supervised methods.
We propose using the pre-trained
VFM with the MLA mechanism to enhance FPN for better
feature extraction, especially in reflective and texture-less regions. Additionally, we present feature-level stereo contrastive
loss and image-level disparity difference loss to improve the
model’s robustness and context-awareness in dealing with
illumination changes and occlusions. Experimental results on
multiple benchmarks validate the effectiveness of the proposed
self-supervised framework.

\bibliographystyle{ieeetr}
\bibliography{mybib}

\end{document}